\documentclass[10pt,journal,compsoc]{IEEEtran}
\usepackage{orcidlink}
\usepackage{cite}
\usepackage{amsmath,amssymb,amsfonts}
\usepackage[ruled]{algorithm2e}
\usepackage{multicol}
\usepackage{multirow}
\usepackage{graphicx}
\usepackage{textcomp}
\usepackage{xcolor}
\usepackage{subcaption}
\usepackage{listings}
\usepackage{setspace}
\usepackage{flushend}
\colorlet{punct}{red!60!black}
\definecolor{background}{HTML}{EEEEEE}
\definecolor{delim}{RGB}{20,105,176}
\colorlet{numb}{magenta!60!black}
\lstdefinelanguage{json}{
    basicstyle=\scriptsize\ttfamily,
    numbers=right,
    numberstyle=\scriptsize,
    stepnumber=1,
    numbersep=1pt,
    captionpos=b, 
    showstringspaces=false,
    breaklines=true,
    frame=lines,
    backgroundcolor=\color{background},
    literate=
     *{0}{{{\color{numb}0}}}{1}
      {1}{{{\color{numb}1}}}{1}
      {2}{{{\color{numb}2}}}{1}
      {3}{{{\color{numb}3}}}{1}
      {4}{{{\color{numb}4}}}{1}
      {5}{{{\color{numb}5}}}{1}
      {6}{{{\color{numb}6}}}{1}
      {7}{{{\color{numb}7}}}{1}
      {8}{{{\color{numb}8}}}{1}
      {9}{{{\color{numb}9}}}{1}
      {:}{{{\color{punct}{:}}}}{1}
      {,}{{{\color{punct}{,}}}}{1}
      {\{}{{{\color{delim}{\{}}}}{1}
      {\}}{{{\color{delim}{\}}}}}{1}
      {[}{{{\color{delim}{[}}}}{1}
      {]}{{{\color{delim}{]}}}}{1},
}
\def\BibTeX{{\rm B\kern-.05em{\sc i\kern-.025em b}\kern-.08em
    T\kern-.1667em\lower.7ex\hbox{E}\kern-.125emX}}
\SetKwComment{Comment}{/* }{ */}
\newlength\mylen
\newcommand\myinput[1]{%
  \settowidth\mylen{\KwData{}}%
  \setlength\hangindent{\mylen}%
  \hspace*{\mylen}#1\\}

\begin{document}

\title{\textsc{Asyn2F}: An Asynchronous Federated Learning Framework with  Bidirectional Model Aggregation}

\author{Tien-Dung Cao, Nguyen T. Vuong, Thai Q. Le, Hoang V.N. Dao,  Tram Truong-Huu\,\orcidlink{0000-0003-1049-9557}~\IEEEmembership{Senior Member,~IEEE}

\IEEEcompsocitemizethanks{\IEEEcompsocthanksitem T.-D. Cao, N.T. Vuong, T.Q. Le and H.V.N Dao are with the School of Information Technology, Tan Tao University, Vietnam. E-mail: dung.cao@ttu.edu.vn, nguyen.vuong@std.ttu.edu.vn, thai.le@std.ttu.edu.vn, hoang.dao@std.ttu.edu.vn. \protect
\IEEEcompsocthanksitem T. Truong-Huu is with the Singapore Institute of Technology (SIT), Singapore. He is also with the Institute for Infocomm Research (I$^2$R), Agency for Science, Technology and Research (A*STAR), Singapore. E-mail: truonghuu.tram@singaporetech.edu.sg.}
}

\IEEEtitleabstractindextext{%

\begin{abstract}
In federated learning, the models can be trained synchronously or asynchronously. Many research works have focused on developing an aggregation method for the server to aggregate multiple local models into the global model with improved performance. They ignore the heterogeneity of the training workers, which causes the delay in the training of the local models, leading to the obsolete information issue. In this paper, we design and develop \textsc{Asyn2F}, an \textsc{Asyn}chronous \textsc{F}ederated learning \textsc{F}ramework with bidirectional model aggregation. By bidirectional model aggregation, \textsc{Asyn2F}, on one hand, allows the server to asynchronously aggregate multiple local models and results in a new global model. On the other hand, it allows the training workers to aggregate the new version of the global model into the local model, which is being trained even in the middle of a training epoch. We develop \textsc{Asyn2F} considering the practical implementation requirements such as using cloud services for model storage and message queuing protocols for communications. Extensive experiments with different datasets show that the models trained by \textsc{Asyn2F} achieve higher performance compared to the state-of-the-art techniques. The experiments also demonstrate the effectiveness, practicality, and scalability of \textsc{Asyn2F}, making it ready for deployment in real scenarios.
\end{abstract}
}

\maketitle

\section{Introduction}
\label{sec:intro}

The proliferation of the Internet of Things, mobile devices and edge computing enables the collection of a huge amount of data, usually personal data at a large scale\cite{evans2011}. Learning representation and generalization of this distributed data has been a grand challenge for both academia and industry. Federated learning has been developed to address the above challenge and becomes an effective method enabling the training of a global model on datasets stored across a number of storage sites~\cite{XIA2021100008}. Without moving data out of premises (which is not allowed by regulations in some scenarios~\cite{gdpr2023}) to a centralized server (e.g., cloud), federated learning also addresses the problem of data privacy.

A federated learning system is composed of a server and a number of distributed training workers~\cite{10.5555/2999611.2999748}. The workers perform model training on their local (private) data. The server aggregates the models received from the workers and updates the global model before sending it back to the workers for the next training iteration. Many works in the literature have focused on two challenging issues of federated learning: (i) optimizing the performance of the global model trained in federated learning (e.g., accuracy) to that of the model trained in a centralized manner~\cite{CAO2022102413,phong2019}; and (ii) reducing the communication costs between the workers and server due to the transmission of the global and local models~\cite{fedavg2017,CAO2022102413,10.1109/TC.2020.2995593}. For the former, these works propose various approaches for the server to aggregate the local models and update the global model. For the latter, model pruning or model compression has been proposed to reduce the size of the models exchanged between the server and workers. The above works make an assumption that whenever there is a training request from the server, all the workers are ready and have sufficient computational resources for model training. The server will wait for the responses from all the requested workers before doing model aggregation and updating the global model. We refer this to as \textit{synchronous federated learning}.

With the emergence of machine learning marketplace~\cite{DBLP:journals/corr/abs-2003-01593,dungcao2022} implementing federated learning, many data providers (maybe with limited computational resources) can contribute their data to train models of the market customers without worrying about data privacy. Many training requests could be sent to the same data provider or the computational resources of data providers are heterogeneous, making the above assumption no longer valid. Synchronous federated learning becomes harder to achieve as the server has to unnecessarily wait for a longer time for all the workers to complete their training. Thus, there is a need for a novel approach that allows the server to asynchronously update the global model once a training worker has completed its training epoch, referred to as \textit{asynchronous federated learning}. While there are a few works that consider this challenge~\cite{xie2020asynchronous,even2022asynchronous,Liu2022,Assran_AGPush_2021, Chen2020_ASO_Fed, Wu2022_KAFL, wang2022asyncfeded, Zhang_2023_FedMDS}, they mainly focus on developing an aggregation method for the server to aggregate multiple local models into the global model with an aim to approximate the performance of the global model to its counterpart trained in a centralized manner. The training workers need to finish a training epoch (iteration) before receiving the new global model from the server. Due to the heterogeneity of training workers, the global model could have been updated several times from the local models obtained from other workers, and the local model of a slow worker becomes obsolete, thus wasting time and computational resources.

Yet, less research looks into the practical implementation of a federated learning framework considering the security policies of data providers. With the increasing security threats, most networks close all ports by default (even common ports such as port $22$, which is used for remote access). This forces the implementation of federated learning frameworks (or marketplaces) to use registered (and well-known) services for the communications between the server and workers whether it is control messages or data. This is to say that conventional socket programming using ephemeral ports may not be suitable.

In this paper, we design and develop \textsc{Asyn2F}, an \textsc{Asyn}chronous \textsc{F}ederated learning \textsc{F}ramework with bidirectional model aggregation that addresses the above challenges. We develop two novel aggregation algorithms used for the server and training workers, respectively. On the server side, the aggregation algorithm allows the server to asynchronously aggregate multiple local models into the global model without waiting for all the training workers to submit their local model. On the worker side, the aggregation algorithm allows the workers to aggregate the new version of the global model into the local model, which is being trained even in the middle of a training epoch, thus reducing the impact of the obsolete information issue. The aggregation algorithms take into account the level of obsolete information, the quality of training data, and the performance of local models, represented by the loss value.

We present the design and implementation of \textsc{Asyn2F} in which we employ advanced message queuing protocols for controlling messages exchanged between the server and workers. We use cloud storage (e.g., AWS S3) for storing models and retrieving requests. As these technologies are common, the network traffic will not be blocked by the network firewall of the data provider's infrastructure. We address several practical challenges such as optimizing the communication costs and synchronizing the learning rate among training workers. We implement a training quality monitoring service that allows end-users to closely monitor the performance of the local models obtained from the workers as well as the global model after aggregation by the server. This monitoring service could be very useful to optimize resource utilization and costs: training can be stopped if the model no longer improves its performance. We perform extensive experiments with various datasets to demonstrate the effectiveness of the proposed method and the developed framework (\textsc{Asyn2F}).

The rest of the paper is organized as follows. Section~\ref{sec:related-work} reviews the literature. Section~\ref{sec:design} presents the design and model aggregation algorithms.  Section~\ref{sec:implementation} presents the implementation of \textsc{Asyn2F}. Section~\ref{sec:exp} presents the experiments and analysis of results. Section~\ref{sec:conclusion} concludes the paper.

\section{Related Work}
\label{sec:related-work}

Asynchronous training has been widely used in traditional distributed Stochastic Gradient Descent~\cite{lian2018asynchronous} for the heterogeneous environment to avoid the blocking time (or delay) of the training process caused by the low performance of computing resources/devices, sometimes called stragglers. The latency of the network bandwidth in a heterogeneous environment also makes the synchronous training process less efficient. Recently, several asynchronous algorithms have been proposed to solve the limitation of synchronous federated learning~\cite{xie2020asynchronous,even2022asynchronous,Liu2022,Assran_AGPush_2021,Chen2020_ASO_Fed,Wu2022_KAFL,Zhang_2023_FedMDS} by removing the delay of stragglers due to both computing performance and network bandwidth. Xie \textit{et al.}~\cite{xie2020asynchronous} defined an algorithm that allows the server to update the global model by weighted average immediately whenever receiving a local model from a worker. The server then pushes the newly-updated global model back to the worker for the next training round without waiting for other workers. However, this approach leads to the problem of obsolete information discussed above. Considering a peer-to-peer training platform where there is no central coordinator, the workers communicate with each other like a graph topology, Assran \textit{et al.}~\cite{Assran_AGPush_2021} and Even \textit{et al.}~\cite{even2022asynchronous} proposed to use the asynchronous gossip algorithm~\cite{gossip} to optimize the models. In these works, each worker has one sending buffer and one receiving buffer; the method repeats two steps until reaching the stop condition. In the first step, each worker locally optimizes the model by epochs (or training rounds). In the second step, the worker sends its model to the neighboring workers after completing a training round/epoch. At the same time, it updates its local model using all the models in the receiving buffer before starting a new training round.

Liu \textit{et al.}~\cite{Liu2022} proposed an asynchronous federated learning mechanism, named Fed2A, that focuses on reducing the communication cost without damaging the overall learning performance. In this mechanism, workers can start training and uploading their local models based on their own decisions, e.g., after a predefined number of training epochs. To reduce the communication cost, the workers divide the model into two parts: i) shallow layers (i.e., feature extraction such as convolutional layers), and ii) deep layers (i.e., classification/regression such as fully connected layers). Each part has a different uploading decision. The server also defines its own aggregation condition, e.g., after receiving a predefined number of local models from workers. When the aggregation condition is satisfied, the server combines the time of transmission delay, data size, and difference in weights between the global model and local models to create a new global model, and then broadcast again to the workers for a new training round. Yu \textit{et al.}~\cite{Async-HFL-2023} proposed an asynchronous and hierarchical framework for federated learning which is used for unreliable IoT networks. The authors proposed two levels of asynchronous model aggregation, i.e., at the gateway and cloud server, to reduce the delay of both network and computing devices on the global model since the number of workers is large. 

Chen \textit{et al.}~\cite{Chen2020_ASO_Fed} defined a method for asynchronous online federated learning. At each step, this method first uses attention mechanisms to extract the features from the incoming data of each worker and aggregates them to have the feature representation learning on the server. It then uses this feature vector to update the global model. Adapting FedAvg~\cite{fedavg2017}, Wu and Wang~\cite{Wu2022_KAFL} developed an Asynchronous Federated Learning framework, namely KAFL with two algorithms. The first one is named K-FedAsyn which waits for the first $K$ fastest workers that run FedAvg to update the global model. The difference with the original FedAvg is that the average of local models is merged to the global model by a ratio $\alpha \in (0,1)$. The second one is named M-Step KAFL, which defines a buffer with a limited size of $M$. Whenever the server receives a local model from a worker, it assigns this local model as the global model, denoted as $W^g_t$. When the buffer is full, the average of local models in the buffer is computed and then merged into $W^g_t$ by a ratio $\alpha \in (0,1)$. Wang \textit{et al.}~\cite{wang2022asyncfeded} defined an asynchronous federated learning algorithm, in which a worker can join the training process at any time instant and get the latest version of the global model to train on its local data. The worker computes the difference between the downloaded global model and the local model after its training iteration and sends the difference to the server. The server then uses the Euclidean distance to compute the adaptive learning rate to update the global model based on the information received from the worker. Recently, Zhang \textit{et al.}~\cite{Zhang_2023_FedMDS} developed FedMDS, a framework for semi-asynchronous federated learning. Unlike FedAvg, FedMDS classifies workers into different groups. In each group, the workers synchronously send their local model to the server for aggregation, but the aggregated model of each group is asynchronously updated by the server. Miao \textit{et al.}~\cite{10215505} recently proposed an asynchronous federated learning algorithm using time-weighted and stale model aggregation. The algorithm sets a duration threshold to classify the workers into online workers or stragglers, then uses the delay to compute the contribution weightage of local models into the global model.

All the above works propose methods for aggregation of the global model at the server but not at the workers, thus raising the issue of obsolete information for the workers. They also do not consider a practical implementation of the framework in which multiple practical implementation requirements need to be addressed. 
\section{Framework Design and Model Aggregation Algorithms}
\label{sec:design}

In this section, we present our proposed framework. We first provide a brief overview of framework architecture. We then describe the aggregation algorithms that the server and the workers use for aggregating and updating models. 

\subsection{General Framework Architecture}
\label{subsec:architecture}

\begin{figure}[t]
    \centering
    \includegraphics[width=0.48\textwidth]{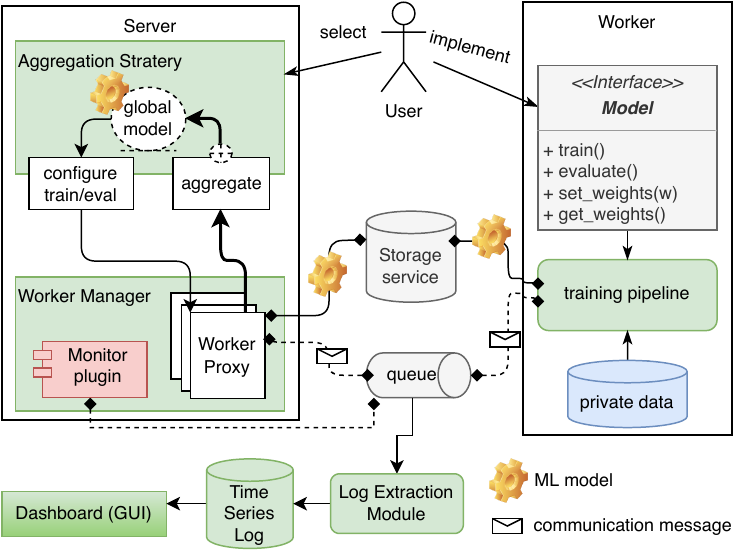}
    \caption{General Architecture of \textsc{Asyn2F}.}
    \label{fig:asynfed_arch}
    %\vspace{-2.5ex}
\end{figure}

The general architecture of \textsc{Asyn2F} is depicted in Fig.~\ref{fig:asynfed_arch}. The server side consists of two main components. The \texttt{Aggregation Strategy} component includes various model aggregation algorithms. We aim to design our framework as generic as possible so that it can integrate not only our aggregation algorithm but also other algorithms such as the existing ones. The \texttt{Worker Manager} component manages all the workers of the framework. The \texttt{Worker Proxy} helps the server communicate with the workers via advanced message queuing protocols for exchanging control messages, e.g., notification of a new global model or local model submissions. The \texttt{Storage Service} is a third-party component (i.e., a cloud service) used for storing the local models obtained from the workers and the global model aggregated by the server. On the worker side, we design a \texttt{Model} interface that allows users to implement the model using their preferred libraries. The interface also includes functions for the workers to aggregate a new global model into the local model during the training process. Given a defined model, the training pipeline includes functions to retrieve training data and invoke the model training on computing resources. The training pipeline also includes a monitoring module that reports the performance of the local models as well as the training status to a monitoring dashboard for real-time visualization. Via real-time monitoring, users can control the training process such as stopping the training process at a specific worker or the entire training process at the server when achieving a desired performance.

\subsection{Asynchronous Federated Learning Algorithms}
\subsubsection{Algorithm Overview and Illustration}

The training process of a model with \textsc{Asyn2F} is presented in Algorithm~\ref{alg:training}, which is executed by the main thread at the server. The server initializes a global model, denoted as $W_0^\text{global}$. The server notifies all the workers joining the training to download the global model and start their training (i.e., execute \texttt{notify\_workers()}). As discussed in the framework architecture, let $Q$ denote the queue that contains the local models of the participating workers. While all the workers execute their local training process, the server waits for local models sent from the workers and keeps checking the aggregation condition (i.e., execute \texttt{check\_aggregation\_cond}($Q$)). If the aggregation condition is satisfied, the server performs model aggregation and produces a new global model, versioned as $W_i^\text{global}$, $i=1,2,\dots$. The aggregation can be triggered periodically or based on the size of the queue that contains notifications received from the workers. For instance, the server updates the global model whenever there is a new local model submitted. The server can also update the global model when there are $k$ models received from $k$ workers (i.e., the case of FedAvg~\cite{fedavg2017}). If we set $k$ to the number of workers joining the training process, the training becomes entirely synchronous training in which the server waits for all the workers to submit their local model before doing aggregation.

The training will be terminated once the stopping condition is satisfied. Conventionally, the training is terminated when it reaches a pre-defined number of training epochs. In practice, users may consider the trade-off between the model performance and the training cost (e.g., time or computing resources of the workers). Thus, we incorporate two other different stopping conditions including: (i) maximum training duration, and (ii) desired performance of the model.

\setlength{\textfloatsep}{0.4cm} 
\begin{algorithm}[t]
\caption{Training process}
\label{alg:training}
\KwData{$W_p$: pre-trained global model;\\
$\mspace{40mu}$ $Q$: queue contains the local models of workers;}
\BlankLine
$W_0^\text{global} \gets W_p$; /*initial global model*/\\
\texttt{notify\_workers()}\;
$i \gets 0$\;
$training \gets True$\;
$aggregation \gets False$\;
\While{training = True}{
  \uIf{$aggregation = True$}{
    $i \gets i + 1$; /* version of global model*/
    $W_i^\text{global} \gets$ \texttt{global\_aggregation()}\;
    \uIf{\texttt{check\_stop\_condition}($W_i^\text{global}$)}{
    $training \gets False$\;
    }
    \Else{
    \texttt{notify\_workers()}\;
    $aggregation \gets False$\;
    }
    }
  \Else{
    \texttt{sleep(period)}\;
    \If {\texttt{check\_aggregation\_cond}($Q$)}{
        $aggregation \gets True$\;
    }
    }
}
\end{algorithm}

\begin{figure*}[t]
    \centering
    \includegraphics[width=0.98\textwidth]{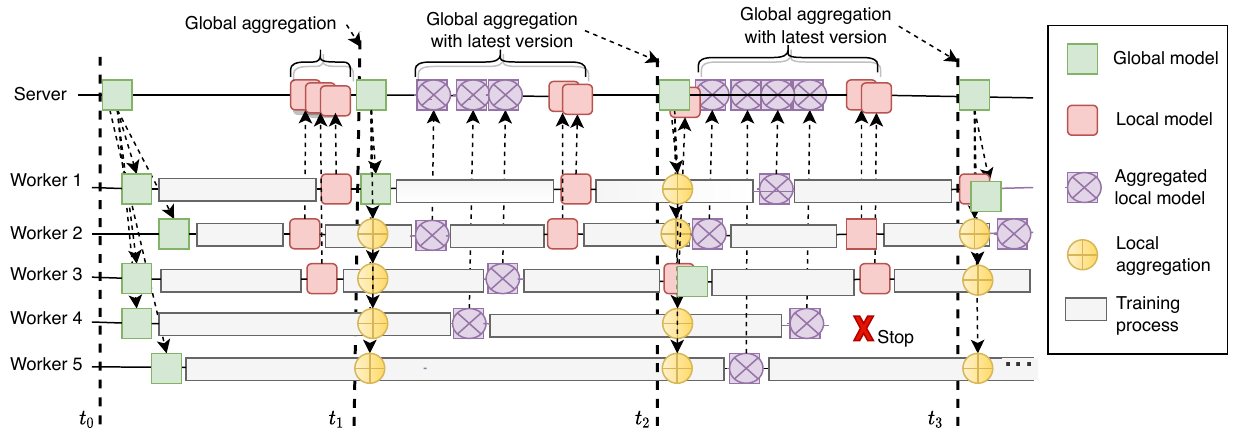}
    \caption{Illustration of asynchronous updates with \textsc{Asyn2F}.}
    \label{fig:asynfed_illustration}
    \vspace{-2ex}
\end{figure*}

The training process is illustrated in Fig.~\ref{fig:asynfed_illustration}. The five workers receive the (initialized) global model (illustrated by light green squares) and start the training process at different time instants. Combined with the heterogeneity in computing resources and the size of training datasets, the workers complete their training epoch and send the local model (represented by pink squares) to the server at different time instants. For instance, Worker 2 finishes its training epoch very early compared to Worker 1 and Worker 3 while Worker 4 spends a longer time to finish its training epoch. Without waiting for the new global model, those workers completing their training epoch early will immediately start a new training epoch with their current local model (e.g., Worker 2 and Worker 3). This significantly reduces the idle time of workers and improves resource utilization. Given that the aggregation condition is satisfied (periodically triggered at time $t_1$, $t_2$, and $t_3$ as shown in Fig.~\ref{fig:asynfed_illustration}), the server performs global model aggregation and creates a new version of the global model. The server will then notify all the workers to download the updated global model. In Fig.~\ref{fig:asynfed_illustration}, Worker 1 just finished its training epoch, thus it will take the updated global model and start the next training epoch. Worker 2 and Worker 3 started the subsequent training epoch earlier, thus requiring a local model aggregation to accommodate the updated global model to the local model that is currently being trained. The local model aggregation process is denoted as yellow circles. The local models obtained from the local model aggregation process (denoted as purple circles in Fig.~\ref{fig:asynfed_illustration}) at the workers will also be sent to the server for the next global aggregation. In this way, our framework avoids the problem of obsolete information at the worker that takes a longer time for a training epoch while the global model has been updated many times by the local models obtained from other workers that complete the training faster.

\subsubsection{Global Model Aggregation at Server}

\setlength{\textfloatsep}{0.3cm} 
\begin{algorithm}[t]
\caption{\texttt{global\_aggregation()}}
\label{alg:server_side}
\KwData{$Q^\text{data}_k$: data quality of dataset $D_k$;\\
$\mspace{40mu}$ $s_k$: size of dataset $D_k$;\\
$\mspace{40mu}$ $Q$: queue contains the local models of workers;\\
$\mspace{40mu}$ $i$: Version index of the global model;
}

\BlankLine
\ForEach{$W_k^\text{local} \in Q$}{
    $\alpha_k \gets \displaystyle\frac{Q_k^\text{data} \times s_k}{L_k\times(i-i_k)}$; 
}    
$\alpha_k = \displaystyle\frac{\alpha_k}{\displaystyle\sum_{k|W_k^\text{local} \in Q} \alpha_k}$; /* normalization*/
$W_i^\text{global} \gets \sum \alpha_k W_k^\text{local}$; 

\Return $W_i^\text{global}$;
\end{algorithm}

Given that the aggregation condition is satisfied, the server starts the global model aggregation process shown in Algorithm~\ref{alg:server_side}. We assume that the server has the metadata of the private datasets of the workers such as the size and quality of the dataset.
Worker $k$ has a private dataset $D_k$, which has its quality denoted by $Q^\text{data}_k$, and the number of data samples is denoted as $s_k$. It is to be noted that there exist many tools for determining the quality of data and the server can rely on them to request the report of data quality from the workers. Furthermore, metrics for the quality of data are different for different applications. There are many metrics that have been discussed extensively in the literature, such as class overlap, feature relevance, timeliness, completeness, and duplicates explained in detail in~\cite{ibm_quality_data_metrics}, which can be used to estimate the quality of data. Technically, the implementation of data evaluation can be diverse~\cite{10.1145/1541880.1541883}.

For a local model received from a worker (denoted as $W_k^\text{local}$ for the local model received from worker $k$), the algorithm computes its contribution to the global model, which depends on the quality of data of the worker, the size of the dataset (i.e., the number of samples), the value of the loss function that is used in the training at the worker and the delay of the local model compared to the global model. The contribution ratio is computed as follows:
\begin{equation}
    \alpha_k = \displaystyle\frac{Q_k^\text{data} \times s_k}{L_k\times(i-i_k)}
    \label{eq:contrate}
\end{equation}
where $Q_k^\text{data}$ and $s_k$ are the quality and size of the dataset of worker $k$. $L_k$ is the value of the loss function obtained after the training at worker $k$. $i_k$ is the version index of the global model that worker $k$ has downloaded and used in its training process, and $i$ is the version index of the resulting global model after aggregation as shown in Algorithm~\ref{alg:training}. Thus, $i-i_k$ is the version delay at worker $k$ in incorporating the newly-released global model in its training. For instance, $i-i_k = 1$ means that worker $k$ has used the latest version of the global model in its training. Thus, the rationale behind Eq.~\eqref{eq:contrate} is that if the value of the loss function and the version delay is large, the contribution of the local model obtained from the worker to the global model should be small. This allows the training to avoid the problem of incorporating obsolete local models received from slow workers into the global model that has been updated by other faster workers. The contribution of each local model will be normalized before being used in updating the global model as shown in the last step of Algorithm~\ref{alg:server_side}. 

It is to be noted that if a fast worker submits multiple versions of its local model, the latest version will be considered for the global model aggregation. As shown in Fig.~\ref{fig:asynfed_illustration}, before time $t_2$ has submitted two versions of its local model. The first one represented by the purple color square is the result of a local aggregation process when the server releases the global model aggregated at time $t_1$. The second one is the result of the subsequent training epoch without aggregating with any new global model.

\subsubsection{Local Model Aggregation at Workers}

Given a new version of the global model, the workers use Algorithm~\ref{alg:client_side} to incorporate the global model into their training immediately. In the mini-batch training approach, the training process at the workers takes a batch of data from the dataset and trains the model using stochastic gradient descent. After each batch, the workers check if there is a new version of the global model released by the server. If so, the workers will download the global model and aggregate it into the local model before starting a new batch. 

Without loss of generality, we assume that worker $k$ is updating the latest global model denoted as $W_i^\text{global}$ where $i$ is the version index of the model to the local model that is being trained. $W_i^\text{global}$ has been computed by the server using Algorithm~\ref{alg:server_side} and it is contributed by a number of workers whose average quality of data is $\overline{Q^\text{data}}$. The total number of samples in the datasets of those workers is denoted as $S_i$ and the average loss of the local models contributing to $W_i^\text{global}$ is denoted as $\overline{L}$. We also assume that the local model that is being trained has been initiated from the global model versioned $i_k$ ($W_{i_k}^\text{local}$). This model has been trained on the private dataset of worker $k$ for $j$ data batches.

\setlength{\textfloatsep}{0.4cm} 
\begin{algorithm}[t]
\caption{Model Aggregation at Workers}
\label{alg:client_side}
\KwData{$Q^\text{data}_k$: data quality of dataset $D_k$;\\
$\mspace{40mu}$ $s_k$: size of dataset $D_k$;\\
$\mspace{40mu}$ $W_i^\text{global}$: The global model;\\
\myinput{$\overline{Q^\text{data}}$: Average QoD of the workers contributing in $W_i^\text{global}$;}
\myinput{$S_i$: Total size of data of the workers contributing in $W_i^\text{global}$;}
\myinput{$\overline{L}$: Average loss of the local models contributing in $W_i^\text{global}$;}
\myinput{$W_{i_k,j}^\text{local}$: Local model being trained at batch $j$;}
$\mspace{40mu}$ $i$: Version of the lastest global model;\\
$\mspace{40mu}$ $i_k$: Version of the global model used by $k$;
}
\BlankLine
$\alpha_k^\text{local} \gets  \beta\displaystyle\frac{Q_k^\text{data} \times s_k}{Q_k^\text{data} \times s_k + \overline{Q^\text{data}} \times S_i} + (1-\beta) \frac{\overline{L}}{L_{i_k,j} + \overline{L}}$\;
\ForEach{$W_i^\text{global}(w)$}{
    $e^\text{local}(w) \gets W_{i_k,j}^\text{local}(w) - W_{i_k,j-1}^\text{local}(w)$\;
    $e^\text{global}(w) \gets W_i^\text{global}(w) - W_{i_k,j}^\text{local}(w)$\;
    \uIf{$e^\text{local}(w) \times e^\text{global}(w) > 0$}{
    $W_{i_k,j}^\text{local}(w) \gets W_i^\text{global}(w)$\;
    }
    \Else{
    $W_{i_k,j}^\text{local}(w) \gets (1-\alpha_k^\text{local}) \times W_i^\text{global}(w) + \alpha_k^\text{local} \times W_{i_k,j}^\text{local}(w)$\; 
    }
    }
\Return $W_{i_k,j}^\text{local}$;    
\end{algorithm}

The algorithm computes a weighted parameter to determine the contribution of the latest global model and that of the local model at batch $j$ to the new version of the local model. The formula used to compute this parameter is defined in Eq.~\eqref{eq:alpha_at_client} where $\beta \in (0,1)$ is a hyper-parameter that defines the importance of the quality of data or the performance of the models represented by the loss. 
\begin{equation}
\label{eq:alpha_at_client}
    \alpha_k^\text{local} =  \beta\displaystyle\frac{Q_k^\text{data} \times s_k}{Q_k^\text{data} \times s_k + \overline{Q^\text{data}} \times S_i} + (1-\beta) \frac{\overline{L}}{L_{i_k,j} + \overline{L}}.
\end{equation}
The above formula takes into account two important factors which are the quality of data and the model performance represented by the loss. Rationally, the higher the quality of data at worker $k$, the higher the contribution of the current local model ($W_{i_k,j}^\text{local}$) to the updated version. If the loss of the current local model is low, it will also have a higher contribution to the updated version of the local model.

For every parameter (weight) of the model, the algorithm calculates the movement distance and direction between two data batches and between the latest global model and the local model. For instance, between two data batches, the movement distance of weight $w$ is calculated as $e^\text{local}(w) = W_{i_k,j}^\text{local}(w) - W_{i_k,j-1}^\text{local}(w)$. If $e^\text{local}(w)$ is greater than zero, it means that $w$ increases from batch $j-1$ to batch $j$ and vice versa. Similarly, the movement distance and direction between the latest global version and the local version is calculated as $e^\text{global}(w) = W_i^\text{global}(w) - W_{i_k,j}^\text{local}(w)$. Given the values of $e^\text{local}(w)$ and $e^\text{global}(w)$, weight $w$ of the local model at batch $j$ (i.e., $W_{i_k,j}^\text{local}(w)$) will be updated accordingly. If $e^\text{local}(w)$ and $e^\text{global}(w)$ have the same sign (i.e., $e^\text{local}(w) \times e^\text{global}(w) > 0$ or both the local model and global model move in the same direction), $W_{i_k,j}^\text{local}(w)$ will be updated directly to $W_{i}^\text{global}(w)$. Otherwise, $W_{i_k,j}^\text{local}(w)$ is computed based on the contribution from both the global model and the current version of the local model as follows: 
\begin{equation}
    W_{i_k,j}^\text{local}(w) = (1-\alpha_k^\text{local}) \times W_i^\text{global}(w) + \alpha_k^\text{local} \times W_{i_k,j}^\text{local}(w)
    \label{eq:merging}
\end{equation}
where $\alpha_k^\text{local}$ is defined in Eq.~\eqref{eq:alpha_at_client}.

The updated local model returned by Algorithm~\ref{alg:client_side} ($W_{i_k,j}^\text{local}$) continues being trained on the next data batches at worker $k$ until completing the training epoch and notifies the server the availability of a new local model.

\subsection{Privacy Preservation Analysis}

Data privacy is the most important factor that motivates the development of federated learning. We prove that the framework developed in our work protects the data of every worker participating in the training process of a model. We consider a threat model in which there is a total of $N-1$ workers that collude with each other to attack the remaining worker with the objective of learning the data of the victim. During each training epoch, there are two scenarios that can happen to the local model trained by a worker. We analyze the data privacy issue of these two scenarios below.

\begin{enumerate}
    \item The local model is completely trained with all the data batches of the worker without being aggregated with any version of the global model. For instance, a worker has a small dataset and the training time for each epoch is too short compared to that of other workers. By applying mini-batch gradient descent with private training parameters, it has been proved in~\cite{phong2019,CAO2022102413} that the honest-but-curious server and/or $N-1$ workers have no information on the private dataset of the victim. 
    
    \item Local model aggregation has happened during a training epoch, i.e., Algorithm~\ref{alg:client_side} is executed one or several times in a training epoch. Without loss of generality, we assume that the victim worker is at the $t^\text{th}$ training epoch. This implies that the dataset of the worker is protected by non-linear equations in $t-1$ previous training epochs. The local model aggregation process presented in Algorithm~\ref{alg:client_side} adds an additional non-linear equation shown in Eq.~\eqref{eq:merging}. This further enhances privacy-preserving property, thus protecting the private datasets more securely. 
\end{enumerate}

We note that as the local datasets are protected by our training approach, the man-in-the-middle threat model will not affect the privacy-preserving property of our framework. Man-in-the-middle adversaries can sniff the training results (local models) during the exchanges between workers and the server but they cannot infer the data even if the models are transferred in plain text without any encryption techniques. However, this may not happen as network traffic nowadays is usually encrypted with Transport Layer Security (TLS) protocol.

\section{Framework Implementation}
\label{sec:implementation}

In this section, we first present the essential requirements of an asynchronous federated learning framework and then we present the technologies and implementation details of our framework, \textsc{Asyn2F}.

\subsection{Framework Functional Requirements}
\label{sec:framework_requirement}

We consider a practical context of federated learning such as machine learning marketplace~\cite{dungcao2022} where users whether they are data providers equipped with computing resources for model training or end-users who request models. In this context, the data providers can join the training framework at any time. Their computing resources (the workers) are heterogeneous and may fail in the middle of the training process (e.g., due to network issues or other reasons). We believe that the following requirements are essential for the success of the federated learning framework.

\subsubsection{Monitoring Support (R1)}  As the workers can join in and leave the federated learning framework at any time, a monitoring dashboard that supports real-time visualization of the model performance and/or loss value of each worker is necessary. Due to asynchronous communications between the server and workers, as well as the different sizes of datasets, a training epoch may take a longer time for a particular worker than the other, this monitoring dashboard should also report to end-users the current status of the workers and the server about training time and joined in/out timestamp. This helps end-users to stop training when the trained model attains a desired performance or stop a worker if the contribution of the local model is no longer needed. This feature is useful especially when model training incurs a cost, e.g., the cost of computing and storage resources, communication cost, and the cost of training data. 

\subsubsection{Testing Worker (R2)} To provide real-time performance monitoring, a testing worker is needed as the server usually does not have a test dataset. This testing worker can belong to the user who requests the model such that the obtained model attains the desired performance of the requester. 
   
\subsubsection{Threshold for Model Exchange (R3)} The communication cost (either time or monetary cost) is a critical problem of federated learning if models are sent back and forth between the server and the workers frequently. To reduce the impact of this problem, the federated learning framework should allow the workers to train the local model independently using their data until reaching a desired performance or a required number of iterations (i.e., epochs), pre-defined as a training parameter. If the local model of a worker attains this threshold, the worker can start exchanging its local model with the server for global aggregation. This is particularly practical, especially for the models that are trained from scratch and normally have low performance in the first few epochs.
    
\subsubsection{Synchronizing Learning Rate among Workers (R4)} Due to the heterogeneity of computing resources and the size of the training dataset of each worker, the training time for an epoch is different among workers. Since the learning rate is decayed depending on the number of epochs performed, the above issue leads to the fact that the local models are trained at the workers with different learning rates. Those local models when aggregated in the global model affect the stability and convergence speed of the global model. To address this issue, the decay function of each worker must be updated according to the number of training epochs of the global model.

\subsection{Implementation Details}

\subsubsection{Sequence Diagram and Communication Messages} 
\begin{figure}[t]
    \centering
    \includegraphics[width=0.5\textwidth]{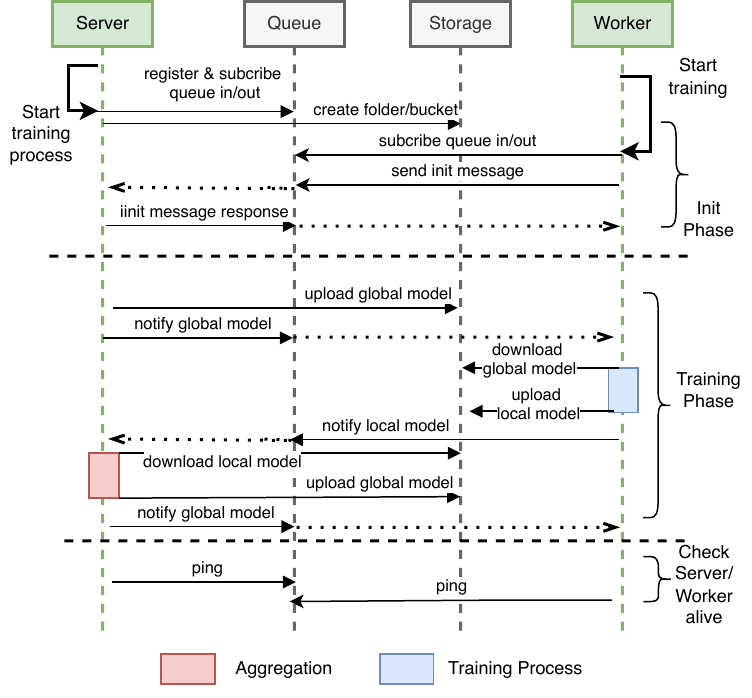}
    \caption{Sequence Diagram between Server and Workers.}
    \label{fig:sequence_diagram}
    %\vspace{-2ex}
\end{figure}

As shown in Fig.~\ref{fig:asynfed_arch}, there are 4  components: server, worker, storage service, and queue in our federated learning framework. In Fig.~\ref{fig:sequence_diagram}, we present the sequence diagram showing the interaction among components during the entire process of model training. The server starts the training process by creating a bucket on the storage service for model exchange and registers the queue channel to publish/subscribe the messages from/to the workers.

As discussed earlier, the workers can join the platform and start the training process at any time. When a worker joins, it first registers the queue for message publishing/subscribing, and once the queue registration is successful, it sends an init message to the server via the queue. The format of the init message of the worker is presented in Listing~\ref{lst:workerinitmessage}. With this message, the worker informs the server about its role in the framework (i.e., trainer or tester based on the requirement \textbf{R2}), its hardware configuration, and metadata of the dataset (i.e., size/number of samples, quality of data). When the server receives the init message from a worker, it creates a WorkerProxy instance and responds to the worker with necessary information packaged in a JSON message presented in Listing~\ref{lst:server_response_message}. This message includes the following information: i) storage information; ii) the current version of the global model and threshold for model exchange (based on the requirement \textbf{R3}); and iii) the aggregation strategy at the server. It is to be noted that the developed framework supports not only the model aggregation approach proposed in our work but also other existing approaches such as FedAvg~\cite{fedavg2017} and M-Step KAFL~\cite{Wu2022_KAFL}. 

\begin{lstlisting}[language=json, caption={Init Message of Workers to Server.},label={lst:workerinitmessage},basicstyle=\ttfamily\footnotesize, showstringspaces=false,  frame=single, breaklines=true, numbers=none]
{
  "headers": {
    "message_type":"WORKER_INIT",
    "worker_id":"id001",
    "session_id":"d53f4850-3452-4224-b874-...",
    "timestamp":"2023-07-15 12:33:56"},
  "content": {
    "role":"trainer|tester",
    "system_info":{
        "cpu":"x86_64",
        "gpu":"NVIDIA GeForce GTX 1080 Ti",
        "ram":"16Gb",
        "disk":"80GB"},
    "data_description":{
        "size":123,
        "qod": 0.95}
    }
}
\end{lstlisting}

\begin{lstlisting}[language=json, caption={Response of Server to Init Message of Workers.},label={lst:server_response_message},basicstyle= \ttfamily\footnotesize, showstringspaces=false,  frame=single, breaklines=true, numbers=none]
{
  "headers": {
    "message_type":"SERVER_INIT_RESP",
    "server_id":"test_cifar10_id01",
    "session_id":"session of worker",
    "timestamp":"2023-07-15 12:34:05"},
  "content": {
    "model_info":{
        "url":"global-models/cifar10_5w_v2.pkl",
        "name":"cifar10_5w",
        "version":2,
        "exchange_at":{
            "performance": 0.9,
            "epoch": 100}
        },
    "storage_info":{
        "access_key":"",
        "secret_key": "",
        "bucket_name":"cifar10_5w",
        "region_name":"asia-southeast-2"},
    "strategy": "asyn2f"
    }
}
\end{lstlisting}

\begin{lstlisting}[language=json, caption={Notification Message of Server to Workers.},label={lst:server_msg_notify},basicstyle= \ttfamily\footnotesize, showstringspaces=false,  frame=single, breaklines=true, numbers=none]
{
  "headers": {
    "message_type":"SERVER_NOTIFY",
    "server_id":"test_cifar10_id01",
    "timestamp":"2023-07-15 12:39:35"},
  "content": {
    "worker_id":["id001"],
    "global_model":{
        "id":"cifar10_model_id01",
        "version": 1,
        "name":"cifar10_5w",
        "total_data_size":42432
        "avg_qod": 0.89
        "avg_loss": 1.232},
    "learning_rate": 0.01 //optional
}
\end{lstlisting}

\begin{lstlisting}[language=json, caption={Notification Message of Workers to Server.},label={lst:worker_msg_notify},basicstyle= \ttfamily\footnotesize, showstringspaces=false,  frame=single, breaklines=true, numbers=none]
{
  "headers": {
    "message_type":"WORKER_NOTIFY",
    "worker_id":"id001",
    "session_id":"d53f4850-3452-4224-b874-...",
    "timestamp":"2023-07-15 12:39:35"},
  "content": {
    "storage_path":"cifar10_5w/id001",
    "file_name":"epoch_1.pkl",
    "global_version_used": 2
    "performance": 0.8934
    "loss": 1.232
}
\end{lstlisting}

During the training process, whenever the server completes the aggregation of the local models received from the workers, it uploads the new global model to the storage and notifies all the workers about the availability of a new global model by sending the message shown in Listing~\ref{lst:server_msg_notify}. The message contains information on the current version of the global model, the average quality of data, and the average loss of the models that were aggregated in the new version of the global model. This information is needed for the workers to perform local aggregation of the global model to the local model as shown in Algorithm~\ref{alg:client_side}. Upon receiving this message, all the workers will need to download the new version of the global model, perform local model aggregation, and continue the training process. 

Upon completion of a training epoch and the local model attains the performance threshold for contribution to the global model, the worker uploads the local model to the storage server and uses the message shown in Listing~\ref{lst:worker_msg_notify} to notify the server. This message contains information about the performance and loss of the local model, the version of the global model that has been used for training the local model. It is to be noted that during a training epoch of the local model, multiple versions of the global model could have been used at different local model aggregations. When notifying the server, the worker takes the latest version of the global model that has been downloaded to include in the notification message.

\subsubsection{Technologies and Implementation Details}

\textbf{Abstract Class for Model Definition.} To support the training of various deep learning models defined by the users of the framework, we define an abstract class, namely \verb|Model|, which allows users to define their own neural network architecture and implements 3 required methods: \verb|get_weight()|, \verb|set_weight()|, and \verb|train()| - for one mini-batch. An optional method for a tester is also defined, namely \verb|evaluate()| allowing users to implement the test function based on their performance metrics. 

\textbf{Messaging System for Communications}. We used RabbitQM\footnote{\url{https://www.rabbitmq.com/}} in our framework. This avoids the use of socket programming and ephemeral ports, which are usually blocked by security systems, e.g., firewalls. All delivered messages are executed by the \verb|routing_key| concept of RabbitMQ. Each worker has two connections to the queue. The first one asynchronously subscribes to the queue using its own queue name and the same \verb|routing_key| to listen to the messages from the server. The second one is used to asynchronously publish messages to the server. Similarly, the server also uses two connections to the queue for receiving/publishing messages from/to workers.

\textbf{Object Storage.} Our framework supports both S3\footnote{\url{https://aws.amazon.com/s3/}} and Minio\footnote{\url{https://min.io/}}. In addition to the two main functions that are uploading and downloading models, our framework also supports several additional functions such as the creation of buckets, and deletion of models that are no longer needed. This is useful in the case of S3 as it is chargeable.   

\textbf{Monitoring and Report.} To meet requirement \textbf{R1}, the monitoring service is implemented by subscribing to messages from workers, the tester, and the server. From these messages, the service extracts necessary information such as \verb|worker_id|, \verb|timestamp|, \verb|loss value|, and \verb|performance| based on a pre-defined metric, and pushes them to InfluxDB\footnote{\url{https://www.influxdata.com/}} and uses Grafana\footnote{\url{https://grafana.com/}} to visualize the information in real-time during the training process.

\textbf{Synchronizing Learning Rate.} To meet the requirement \textbf{R4}, the decay function is defined by using the version of the global model. Both information (i.e., version of the global model and learning rate) are included in the message that the server uses to notify the workers (see Listing~\ref{lst:server_msg_notify}).

\textbf{Supporting Multiple Model Aggregation Strategies.} In our framework, WorkerProxy keeps all the necessary information of a training worker. It also provides an abstract class namely \verb|Strategy| with an \verb|aggregate()| method and a \verb|train()| method. With this abstract class, apart from the aggregation algorithms presented in previous sections, our framework additionally implemented the existing algorithms such as M-Step KAFL~\cite{Wu2022_KAFL}, AsynFedED~\cite{wang2022asyncfeded} or FedAvg~\cite{fedavg2017}. Respectively, the server also implements the \verb|aggregate()| method for the above techniques. This makes our framework effective for not only asynchronous training techniques but also synchronous ones.

\section{Experiments}
\label{sec:exp}

In this section, we present extensive experiments to demonstrate the effectiveness of our proposed federated learning approach along with the developed training framework. We use 2 datasets for our experiments: CIFAR10~\cite{cifar10} and EMBER~\cite{anderson2018ember}. While CIFAR10 is a benchmarking dataset used by many works in the literature, EMBER is a malware dataset used in malware detection applications. It contains the feature vector of 1 million samples including both malware samples and benign samples. Benign samples are actually goodware and sometimes contain sensitive information (e.g., email attachments), thus rarely being shared. For both datasets, we assume that the quality of data is the same when splitting into multiple subsets. Thus, global model aggregation and local model aggregation are mostly based on the loss values. We perform various experimental scenarios, each being described in the below sections with the details of experimental infrastructure, dataset distribution, model architecture, and analysis of results.

We demonstrate the effectiveness of \textsc{Asyn2F} by comparing its performance with existing techniques including FedAvg~\cite{fedavg2017} and M-Step KAFL~\cite{Wu2022_KAFL}. With the two applications mentioned above, we use accuracy as a performance metric. We also use the number of training epochs to demonstrate the convergence speed of the global model.

\subsection{Peformance with CIFAR10}
\subsubsection{Experiments with a small-size infrastructure}

We set up the experimental infrastructure that is composed of 1 server, 1 testing worker, and 10 training workers. The details of location and computing resources are given in Table~\ref{tab:10w_setup}. All the training techniques (\textsc{Asyn2F}, FedAvg, and M-Step KAFL) use the same infrastructure. 

\begin{figure}[t]
    \centering
    \begin{subfigure}[b]{0.48\textwidth}
         \centering
          \includegraphics[width=\textwidth]{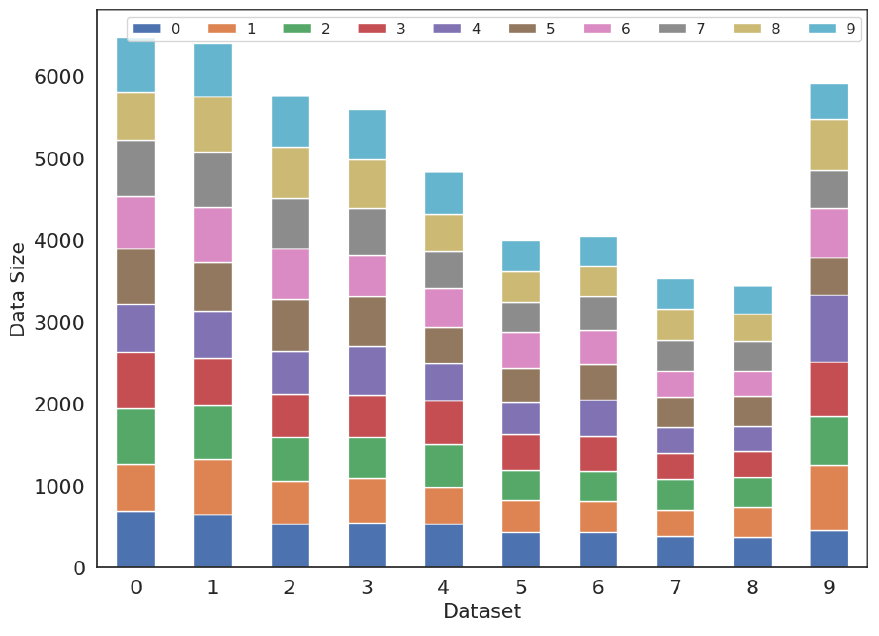}
          \caption{Data size.}
     \end{subfigure}
     
   \begin{subfigure}[b]{0.48\textwidth}
         \centering
          \includegraphics[width=\textwidth]{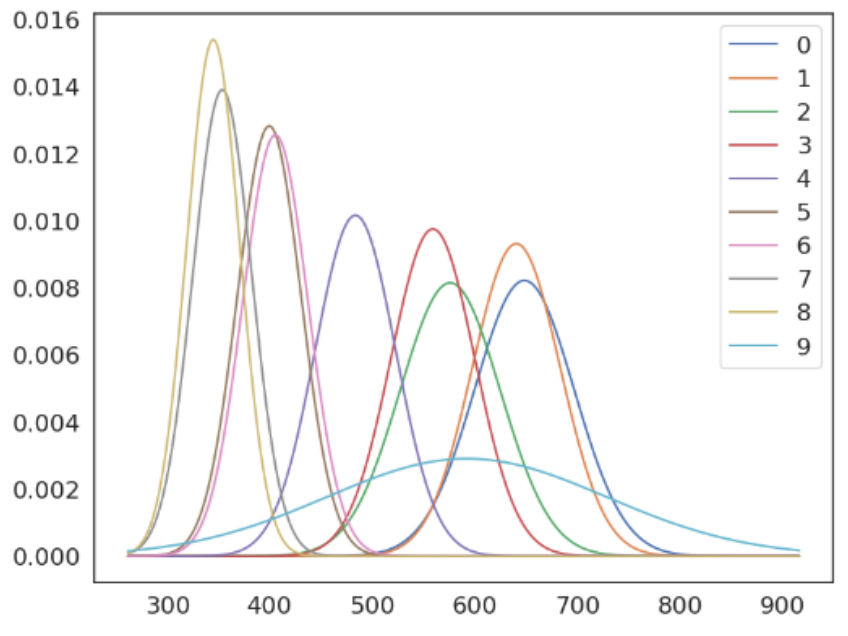}
          \caption{Label distribution.}
     \end{subfigure}
    \caption{Data size and label distribution of 10 sub-datasets (non-overlapping and non-iid).}
    \label{fig:10w_ds}
\end{figure}

\textbf{Data Preparation.}
We use both iid and non-iid random methods to split the CIFAR10 dataset into 10 sub-datasets. When using the iid random method, the sub-datasets are created with two 2 scenarios:
\begin{itemize}
    \item In the first scenario, each sub-dataset is randomly selected from the original dataset by a random ratio from $20\%$ to $40\%$. This leads to the fact that there is an overlap among 10 sub-datasets, and the total data size of 10 sub-datasets is greater than the original size (i.e., $50,000$ samples).
    \item In the second scenario, there is no overlap among the 10 sub-datasets. However, we make sure that every dataset should have samples of all the labels.
\end{itemize} 
When using non-iid random method, all the sub-datasets are non-overlapping. The data distribution of one of the experiments is presented in Fig.~\ref{fig:10w_ds} where the colors represent the labels of the datasets.
\begin{table}[t]
    \centering
    \footnotesize
    \caption{Details of experimental infrastructure with 10 training workers}
    \label{tab:10w_setup}
    \begin{tabular}{|l|r|l|}
    \hline
        \textbf{Role} & \textbf{No. Instances} & \textbf{Description} \\
        \hline
        Storage & 1 & Amazon S3 / CSC - HPC (Finland)\\
        \hline
        Queue & 1 & RabbitMQ of CloudAMQP\\
        \hline
        Server & 1 & CSC - HPC (Finland)\\
        \hline
        Tester & 1 & Google Colab GPU\\
        \hline
        \multirow{3}{*}{Workers} &  3  & 2 GPU Workstations at TTU (Vietnam)\\
        \cline{2-3}
         & 2  & 2 CPU Workstations at TTU (Vietnam)\\
         \cline{2-3}
         & 5  & Google Colab GPU \\
         \hline
    \end{tabular}
    \vspace{-2ex}
\end{table}

\textbf{Model architecture and hyper-parameter setting.}
We use ResNet18 model~\cite{resnet32:2016}, implemented in TensorFlow. We use a stochastic gradient descent optimizer with the \verb|momentum| parameter set to 0.9. We run the experiments with 3 settings of learning rate (LR).
\begin{itemize}
    \item Fixed learning rate: All the workers use the same learning rate for the entire training set to 0.01.
    \item Synchronously-decayed learning rate: The learning rate is decayed along the training process by the server using the cosine decay function with an initial value set to 0.1. The updated learning rate is synchronized with all the workers.
    \item Asynchronously-decayed learning rate: Each worker uses the cosine decay function by its local training steps with an initial value set to 0.1. The updated learning rate is used locally without synchronizing with other workers.
\end{itemize}

For each training approach, we run the training process for 3 hours. Each worker trains 5 iteration rounds of local data as an epoch. For \textsc{Asyn2F}, the global model aggregation process at the server is triggered when there are three local models received from three different workers (always get the latest version). Whereas, M-Step KAFL waits for three models without differentiating workers. The client selection strategy of FedAvg is all clients, i.e., 10 workers. 

\textbf{Analysis of Results.}
In Table~\ref{tab:10w_best_acc_cifar10}, we present the accuracy of the ResNet18 models trained by our training technique and the other two existing techniques. The experimental results show that \textsc{Asyn2F} achieves the best performance compared to the other two existing techniques. The improvement is more significant (up to 2.90\%) in the case of non-overlapping and non-iid sub-datasets, which is usually a practical situation where data providers have different (diverse) datasets. 
Comparing three methods of learning rate setting, we observe that synchronizing the learning rate among workers achieves the best performance. This reflects the requirement (\textbf{R4}) of the training framework discussed in Section~\ref{sec:framework_requirement}. It is worth mentioning that the existing works (FedAvg and M-Step KAFL) do not address this issue in their work. When integrating their training techniques into our framework, we additionally implement their techniques with the three learning rate setting strategies, thus providing insightful performance comparison.

\begin{table}[t]
    \centering
    \footnotesize
    \caption{Accuracy of models trained with 10 workers with different techniques after 3 hours of training}
    \label{tab:10w_best_acc_cifar10}
    \begin{tabular}{|p{2.43cm}|p{1.45cm}|p{1.8cm}|p{1.5cm}|}
    \hline
        \textbf{Strategy} & \textbf{\textsc{Asyn2F}} & \textbf{M-Step KAFL} & \textbf{FedAvg} \\
        \hline
        \multicolumn{4}{|c|}{\textit{Overlapping iid sub-datasets}} \\
        \hline
        Fixed LR=0.01 & $92.86 \pm 0.27$ &  $91.15 \pm 0.43$ & $91.21 \pm 0.74$\\
        \hline
        Sync.-decayed LR & $95.48 \pm 0.09$ & $94.72 \pm 0.18$ & $94.57 \pm 0.13$ \\
        \hline
        Async.-decayed LR& $94.79 \pm 0.16$ & $93.42 \pm 0.50$ & NA\\
        \hline
        \multicolumn{4}{|c|}{\textit{Non-overlapping iid sub-datasets}} \\
         \hline
        Sync.-decayed LR & $94.43 \pm 0.09$  & $92.09 \pm 0.79$ & $93.42 \pm 0.12$ \\
        \hline
        \multicolumn{4}{|c|}{\textit{Non-overlapping, non-iid sub-datasets}} \\
         \hline
        Sync.-decayed LR & $94.17 \pm 0.17$  & $91.27 \pm 0.10$ & $92.52 \pm 0.36$ \\
        \hline
    \end{tabular}
    %\vspace{-2.5ex}
\end{table}

\begin{figure*}[t]
    \centering
    \includegraphics[width=0.98\textwidth]{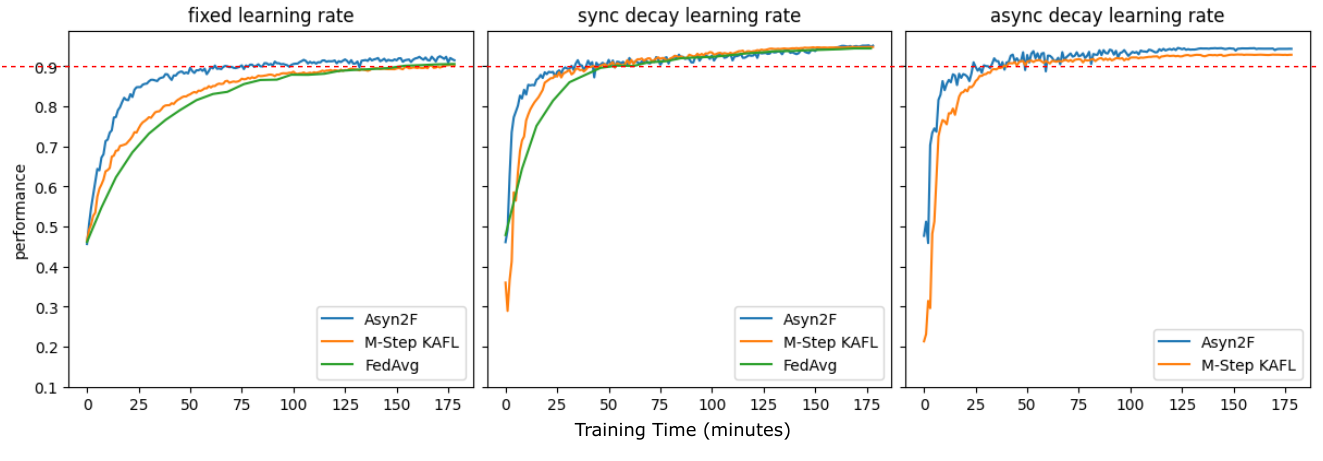}
    \caption{Performance of the global model trained with 10 workers by different techniques (overlapping sub-datasets).}
    \label{fig:10w_acc_3lr_strategy}
    \vspace{-2.5ex}
\end{figure*}

\begin{figure}[t]
    \centering
    \includegraphics[width=0.4\textwidth]{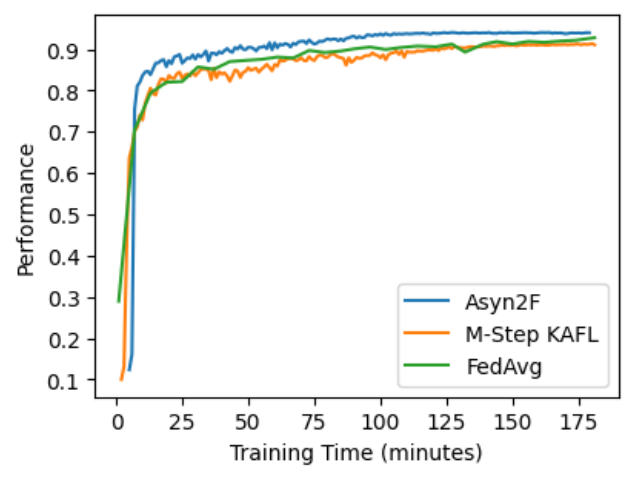}
    \caption{Performance of the global model trained with 10 workers on non-overlapping and non-iid sub-datasets with synchronized learning rate.}
\label{fig:10w_acc_non_overlap_non_iid_cifar}
\vspace{-2.5ex}
\end{figure}

Looking into convergence speed, we present in Fig.~\ref{fig:10w_acc_3lr_strategy} and Fig.~\ref{fig:10w_acc_non_overlap_non_iid_cifar} the performance of the global model obtained by the testing worker over training time in case of overlapping and non-overlapping sub-datasets, respectively. The experimental results show that the global model trained by \textsc{Asyn2F} converges faster compared to the models trained by the other two techniques regardless of the learning rate setting methods. This is explained by the fact that the local model aggregation in \textsc{Asyn2F} enables a faster convergence of local models trained by the workers and avoids the information obsolete issue at the workers. In the long run, all the models will achieve maximum performance (which could not be further improved by any chance). But with a desired performance (e.g., $90\%$ of accuracy), a fast convergence of the model benefits a lot. For instance, the users can stop the training when the global model achieves the desired performance.  From the training cost point of view, this could lead to a lower training cost for both time and monetary cost of computing resources, especially in the context of federated learning marketplaces.

\begin{figure}[t]
    \centering
    \includegraphics[width=0.48\textwidth]{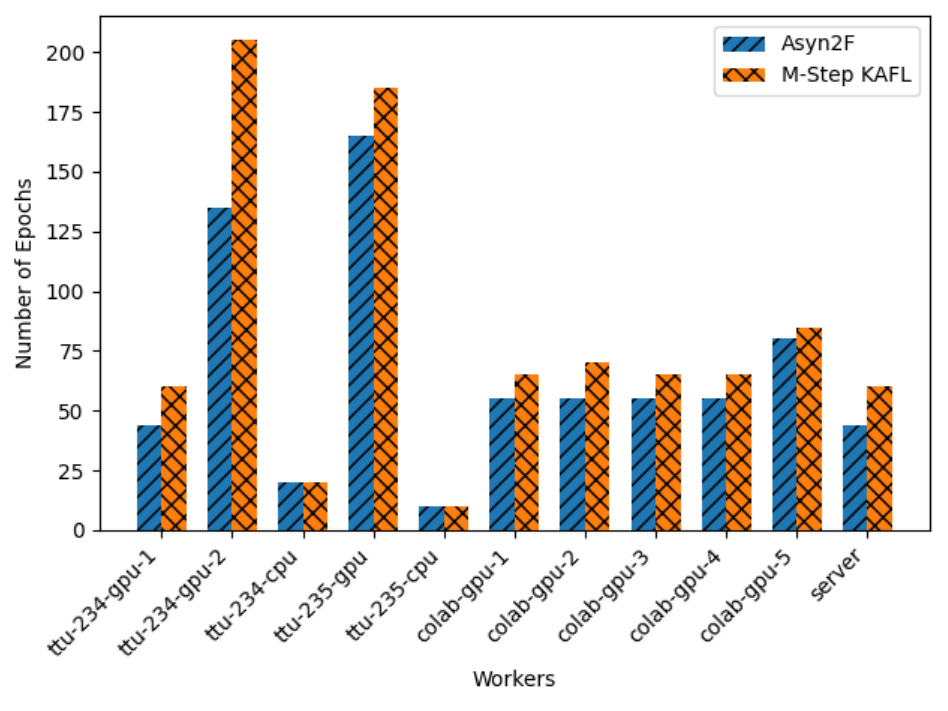}
    \caption{Number of training epochs at workers and server to reach the accuracy target, i.e., 90\%.}
    \label{fig:10w_acc_3lr_strategy_target_90}
    \vspace{-0.5ex}
\end{figure}

A fast convergence of the model also implies a lower communication cost. In Fig.~\ref{fig:10w_acc_3lr_strategy_target_90}, we present the number of training epochs carried out at the workers and the server until the global model achieves $90\%$ of accuracy. The experimental results show that while \textsc{Asyn2F} achieves a faster convergence, all the workers and the server need fewer training epochs compared to M-Step KAFL to converge to the desired performance. This implies that \textsc{Asyn2F} incurs a lower communication cost as the workers and the server need to exchange the obtained models after each training epoch (at the workers) and after each aggregation at the server. The ResNet18 model has a size of $42.7$ MB, which incurs a significant communication cost, especially when we use Amazon S3 as the storage service, which charges monetary cost based on the amount of data uploaded/downloaded from the cloud. It is worth mentioning that the higher the number of models exchanged (stored), the higher the storage cost as well. To reduce the storage cost, those intermediate (local and global) models (obtained prior to the final global model) could be deleted.  

It is to be noted that, in Fig.~\ref{fig:10w_acc_3lr_strategy_target_90},  we do not present the number of training epochs of FedAvg, which requires the same number of training epochs for all the workers and the server (10 epochs to attain $90\%$ of accuracy). However, due to the synchronization among workers, the time duration for each training epoch is very long due to the heterogeneity of computing resources of the workers and their dataset. This explains the reason why FedAvg spends a longer time to converge to the stabilized performance of the model. This demonstrates the practicality of our technique and the implemented framework, which can work very well in a heterogeneous infrastructure including both network connectivity and computing resources. 

\subsubsection{Experiments with larger infrastructures}
In this section, we perform the experiments with a larger training infrastructure including $20$ heterogeneous workers. The details of the infrastructure are presented in Table~\ref{tab:20w_setup}. It is worth mentioning that several workers are running CPU resources, which are much slower than GPU resources when training large deep learning models on large datasets.

\begin{table}[t]
    \centering
    \footnotesize
    \caption{Details of experimental infrastructure with 20 training workers}
    \label{tab:20w_setup}
    \begin{tabular}{|l|r|l|}
    \hline
        \textbf{Role} & \textbf{No. Instances} & \textbf{Description} \\
        \hline
        Storage & 1 & CSC - HPC (Finland)\\
        \hline
        Queue & 1 & RabbitMQ of CloudAMQP\\
        \hline
        Server & 1 & CSC - HPC (Finland)\\
        \hline
        Tester & 1 & Google Colab GPU\\
        \hline
        \multirow{5}{*}{Workers} &  3  & 2 GPU Workstations at TTU (Vietnam)\\
        \cline{2-3}
         & 2  & 2 CPU Workstations at TTU (Vietnam)\\
         \cline{2-3}
         & 2  & 2 Laptop GPU (Home Network)\\
         \cline{2-3}
         & 2  &  GPU Workstation at SIT (Singapore)\\
         \cline{2-3}
         & 11  & Google Colab GPU \\
         \hline
    \end{tabular}
    %\vspace{-0.5ex}
\end{table}

\begin{figure}[t]
    \centering
    \begin{subfigure}[b]{0.48\textwidth}
         \centering
          \includegraphics[width=\textwidth]{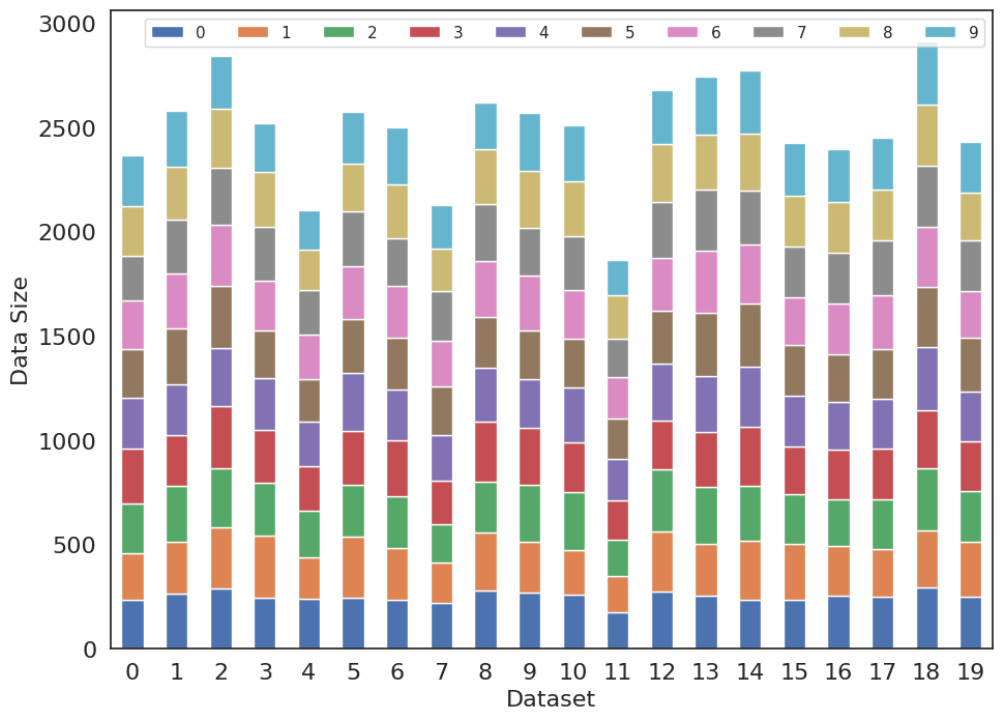}
          \caption{Data size of each subset.}
     \end{subfigure}
     
   \begin{subfigure}[b]{0.48\textwidth}
         \centering
          \includegraphics[width=\textwidth]{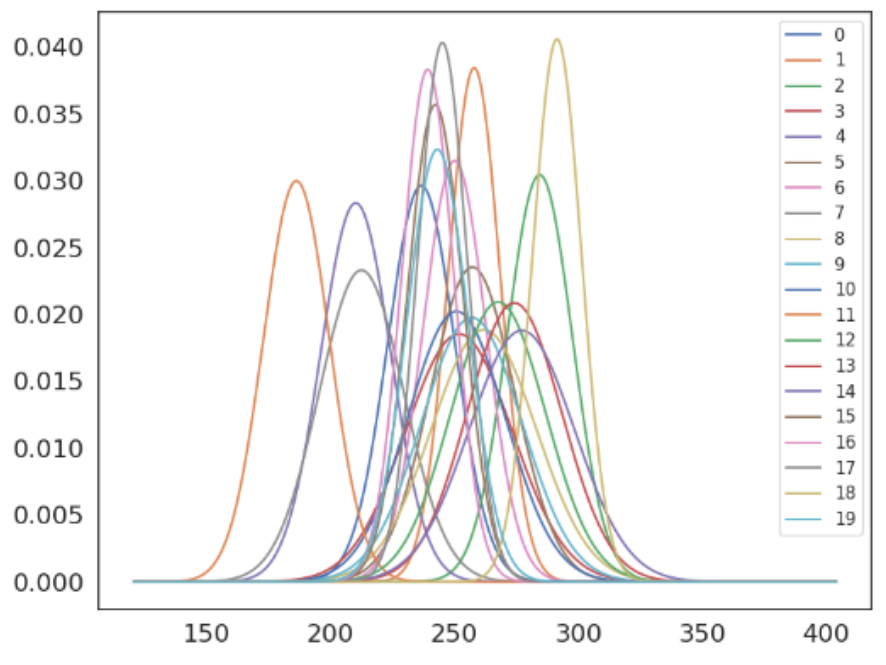}
          \caption{Label distribution.}
     \end{subfigure}
    \caption{Data size and label distribution of 20 sub-datasets.}
    \label{fig:20w_ds}
\end{figure}

\textbf{Data Preparation:} We split the CIFAR10 dataset into 20 sub-datasets with non-iid splitting method. The data size and label distribution of one of the runs are shown in Fig.~\ref{fig:20w_ds}. 

\begin{figure*}[t]
    \centering
    \includegraphics[width=\textwidth]{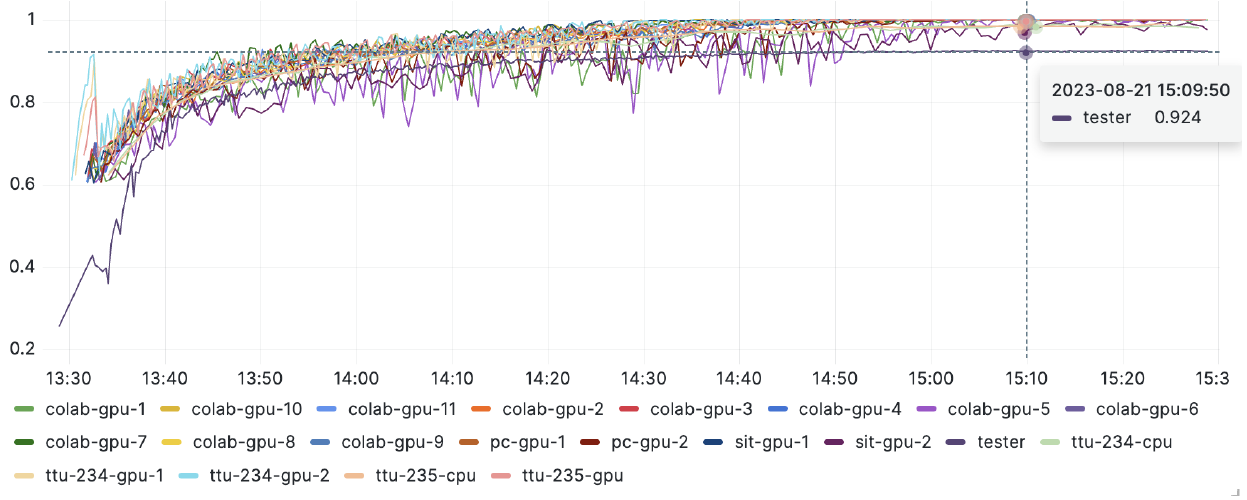}
    \caption{Performance of 20 workers and Tester visualized by Grafana.}
    \label{fig:20w_acc}
    \vspace{-2ex}
\end{figure*}

\textbf{Analysis of Results:} As shown in Table~\ref{tab:20w_best_acc_cifar}, we obtain the same performance behavior of the proposed technique and the existing ones. \textsc{Asyn2F} outperforms both M-Step KAFL and FedAvg. Obviously, compared to the case with fewer workers, there is a performance drop due to the scarce distribution of the same number of training data samples to a larger number of workers. In Fig.~\ref{fig:20w_acc} (captured from the monitoring dashboard of the framework), we present the performance evolution of the local models trained by the workers despite small training datasets. All the local models converge to a stabilized performance thanks to the knowledge obtained from the global model, which is in turn aggregated from other local models. The curve represented by the tester is the performance of the global model that stabilizes after a few hours of training. We also observe an interesting behavior of the framework in this experiment. Due to a large number of training workers and small local training datasets, the workers complete their training epoch very fast, especially the workers with GPU resources. These workers quickly submit their local models to the server for aggregation and continue the next training epoch with the new version of the global model aggregated by the server. The issue of obsolete information does not incur with these workers. Nevertheless, the larger the training framework, the more heterogeneous the workers. Those workers with less powerful computing resources will suffer the consequence of the obsolete information issue, thus significantly benefiting when joining our framework.

It is to be noted that M-Step KAFL needs 10 local models for every global aggregation to achieve an accuracy of $92.10\%$. With a fair comparison with 3 local models for every global aggregation, M-Step KAFL achieves an accuracy of only $83\%$. Further experiments with 7 local models for every global aggregation, M-Step KAFL achieves an accuracy of $89\%$. This makes M-Step KAFL less practical in a heterogeneous federated learning environment. 
\begin{table}[t]
    \centering
    \footnotesize
    \caption{Accuracy of models trained with 20 workers with different techniques on CIFAR10}
    \label{tab:20w_best_acc_cifar}
    \begin{tabular}{|p{2.43cm}|p{1.45cm}|p{1.8cm}|p{1.5cm}|}
    \hline
        \textbf{Strategy} & \textbf{\textsc{Asyn2F}} & \textbf{M-Step KAFL} & \textbf{FedAvg} \\
        \hline
        Sync.-decayed LR & $92.64 \pm 0.05$ & $92.10 \pm 0.12$ & $91.68 \pm 0.18$ \\
        \hline
    \end{tabular}
\end{table}

\subsection{Experiments with EMBER Dataset}

\subsubsection{Experimental settings}
EMBER~\cite{anderson2018ember} is a malware dataset including 1 million samples, each being represented by a vector of 2381 feature values. There are 300000 benign samples (labeled as 0), 300000 malicious samples (labeled as 1) and the remaining 400000 samples are undefined. We remove the 400000 undefined samples and use 600000 labeled samples for our experiments. We split the data into 7 sub-datasets using the non-overlapping and non-iid splitting methods. We adopt MalConv model~\cite{anderson2018ember}, whose detailed parameters are presented in Table~\ref{tab:ember_tf_model} and train it on the training framework with 7 training workers and a tester as shown in Table~\ref{tab:7w_setup_ember}. We note that with only 2381 features representing a sample, the length of the feature vector of the EMBER dataset is much smaller compared to CIFAR10. Even with a deep learning model described in Table~\ref{tab:ember_tf_model} (which is also much smaller than ResNet18 used for CIFAR10), the workers with GPU resources complete a training epoch very fast. This explains why we split the original dataset into only 7 sub-datasets so that the training time of one epoch on a worker will be long enough to demonstrate the heterogeneity of the framework.  

\begin{table}[t]
    \centering
    \caption{Architecture of neural network model}
    \label{tab:ember_tf_model}
    \footnotesize
    \begin{tabular}{|l|l|}
    \hline
        \textbf{Layer} & \textbf{Parameters} \\
        \hline
         Embedding &  in\_dim=261, out\_dim=8\\
        \hline
        \multirow{2}{*}{Conv1D} & filters=128, kernel\_size=15, strides=15, \\
         & use\_bias=True, activation='relu', padding='valid'\\
        \hline
        \multirow{2}{*}{Conv1D} & filters=128, kernel\_size=15, use\_bias=True\\
        & strides=15, activation='sigmoid', padding='valid'\\
        \hline
         Multiply & multiplies (element-wise) layers 2 \& 3\\
        \hline
         \multicolumn{2}{|l|}{GlobalMaxPooling1D} \\
        \hline
         Dense & units=128, activate='relu'\\
        \hline
        Dense & units=1, activate='sigmoid'\\
         \hline
    \end{tabular}
\end{table}

\begin{table}[t]
    \centering
    \footnotesize
    \caption{Details of experimental infrastructure with 7 training workers for EMBER dataset}
    \label{tab:7w_setup_ember}
    \begin{tabular}{|l|c|l|}
    \hline
        \textbf{Role} & \textbf{No. Instances} & \textbf{Description} \\
        \hline
        Queue & 1 & RabbitMQ of CloudAMQP\\
        \hline
        Storage & 1 & CSC - HPC (Finland)\\
        \hline
        Server & 1 & CSC - HPC (Finland)\\
        \hline
        Tester & 1 & Google Colab GPU\\
        \hline
        \multirow{4}{*}{Workers} & 3 & 2 CPU Workstations at TTU (Vietnam)\\
        \cline{2-3}
         & 2 & 2 CPU Desktop at TTU (Vietnam)\\
         \cline{2-3}
         & 1 & 1 GPU Mac M1 (Home - Vietnam)\\
         \cline{2-3}
         & 1 & Google Colab GPU \\
         \hline
    \end{tabular}
\end{table}

\begin{table}[t]
    \centering
    \footnotesize
    \caption{Accuracy of models trained with 7 workers with different techniques after 1 hour of training on EMBER dataset}
    \label{tab:7w_best_acc_ember}
    \begin{tabular}{|p{2.43cm}|p{1.45cm}|p{1.8cm}|p{1.5cm}|}
    \hline
        \textbf{Strategy} & \textbf{\textsc{Asyn2F}} & \textbf{M-Step KAFL} & \textbf{FedAvg} \\
        \hline
        Fixed LR=0.01 & $98.11 \pm 0.11$ &  $97.99 \pm 0.06$ & $98.01 \pm 0.10$\\
        \hline
        Sync.-decayed LR & $98.28 \pm 0.04$ & $98.17 \pm 0.01$ & $97.94 \pm 0.14$ \\
        \hline
        Async.-decayed LR& $98.17 \pm 0.11$ & $98.03 \pm 0.10$ & NA\\
        \hline
    \end{tabular}
\end{table}

\subsubsection{Analysis of Results} The performance reported in Table~\ref{tab:7w_best_acc_ember} demonstrates the effectiveness of the training framework as well as the aggregation algorithms. We obtain similar performance behavior as with CIFAR10 such that \textsc{Asyn2F} has a superior performance compared to the existing ones. The results also confirm the fact that synchronizing the decayed learning rate among training workers results in the best performance.

\section{Conclusion}
\label{sec:conclusion}

In this work, we designed and developed \textsc{Asyn2F} an asynchronous federated learning framework with bidirectional aggregation. The framework addresses the intrinsic characteristics of a distributed computing environment where computing and network resources are heterogeneous combined with diversity in training datasets of data providers. We developed two aggregation algorithms used for the server and the workers, respectively. The global model aggregation algorithm allows the server to aggregate multiple local models to create a new version of the global model. The local model aggregation allows training workers to merge the global model into the local model which is being trained as soon as a new version of the global model is available. This addresses the obsolete information issue and enables a fast convergence of the models. We carried out extensive experiments on a benchmarking dataset (CIFAR10) and a more practical dataset (EMBER) to validate the performance of \textsc{Asyn2F}. The experimental results show the superiority of the proposed framework over the recently developed ones.  The framework demonstrates its practicality with the use of cloud services for its implementation, thus overcoming the security restrictions of enterprises or organizations to protect the privacy of their data.

\section*{Acknowledgements}
The work is supported by Tan Tao University Foundation for Science and Technology Development under Grant No. TTU.RS.22.102.001. The authors wish to acknowledge CSC IT Center for Science, Finland, for cloud resources.
\bibliographystyle{IEEEtran}
\bibliography{ref}

\begin{thebibliography}{10}
\providecommand{\url}[1]{#1}
\csname url@samestyle\endcsname
\providecommand{\newblock}{\relax}
\providecommand{\bibinfo}[2]{#2}
\providecommand{\BIBentrySTDinterwordspacing}{\spaceskip=0pt\relax}
\providecommand{\BIBentryALTinterwordstretchfactor}{4}
\providecommand{\BIBentryALTinterwordspacing}{\spaceskip=\fontdimen2\font plus
\BIBentryALTinterwordstretchfactor\fontdimen3\font minus
  \fontdimen4\font\relax}
\providecommand{\BIBforeignlanguage}[2]{{%
\expandafter\ifx\csname l@#1\endcsname\relax
\typeout{** WARNING: IEEEtran.bst: No hyphenation pattern has been}%
\typeout{** loaded for the language `#1'. Using the pattern for}%
\typeout{** the default language instead.}%
\else
\language=\csname l@#1\endcsname
\fi
#2}}
\providecommand{\BIBdecl}{\relax}
\BIBdecl

\bibitem{evans2011}
D.~Evans, ``{The Internet of Things: How the Next Evolution of the Internet Is
  Changing Everything},'' Apr. 2011, {White Paper, Cisco}.

\bibitem{XIA2021100008}
Q.~Xia, W.~Ye, Z.~Tao, J.~Wu, and Q.~Li, ``A survey of federated learning for
  edge computing: Research problems and solutions,'' \emph{High-Confidence
  Computing}, vol.~1, no.~1, p. 100008, 2021.

\bibitem{gdpr2023}
\BIBentryALTinterwordspacing
EU, ``{European Union’s General Data Protection Regulation (GDPR) },'' Apr.
  2023. [Online]. Available: \url{https://gdpr-info.eu/}
\BIBentrySTDinterwordspacing

\bibitem{10.5555/2999611.2999748}
Q.~Ho \emph{et~al.}, ``{More Effective Distributed ML via a Stale Synchronous
  Parallel Parameter Server},'' in \emph{Proc. NIPS'13}, Lake Tahoe, Nevada,
  2013.

\bibitem{CAO2022102413}
T.-D. Cao, T.~Truong-Huu, H.~Tran, and K.~Tran, ``A federated deep learning
  framework for privacy preservation and communication efficiency,''
  \emph{Journal of Systems Architecture}, vol. 124, 2022.

\bibitem{phong2019}
L.~T. Phong and T.~T. Phuong, ``{Privacy-Preserving Deep Learning via Weight
  Transmission},'' \emph{IEEE Transactions on Information Forensics and
  Security}, vol.~14, no.~11, pp. 3003--3015, Nov. 2019.

\bibitem{fedavg2017}
H.~B. {McMahan} \emph{et~al.}, ``{Communication-Efficient Learning of Deep
  Networks from Decentralized Data},'' in \emph{International Conference on
  Artificial Intelligence and Statistics}, 2017.

\bibitem{10.1109/TC.2020.2995593}
C.~Gong, Y.~Chen, Y.~Lu, T.~Li, C.~Hao, and D.~Chen, ``{VecQ: Minimal Loss DNN
  Model Compression With Vectorized Weight Quantization},'' \emph{IEEE Trans.
  Comput.}, vol.~70, no.~5, May 2021.

\bibitem{DBLP:journals/corr/abs-2003-01593}
A.~Kumar, B.~Finley, T.~Braud, S.~Tarkoma, and P.~Hui, ``Marketplace for {AI}
  models,'' \emph{CoRR}, vol. abs/2003.01593, 2020.

\bibitem{dungcao2022}
T.~Cao, H.-L. Truong, T.~Truong-Huu, and M.-T. Nguyen, ``{Enabling Awareness of
  Quality of Training and Costs in Federated Machine Learning Marketplaces},''
  in \emph{IEEE UCC 2022}, CA, USA, Dec. 2022.

\bibitem{xie2020asynchronous}
C.~Xie, S.~Koyejo, and I.~Gupta, ``Asynchronous federated optimization,'' in
  \emph{12th Annual Workshop on Optimization for Machine Learning}, 2020.

\bibitem{even2022asynchronous}
M.~Even, H.~Hendrikx, and L.~Massouli{\'e}, ``Asynchronous speedup in
  decentralized optimization,'' in \emph{Workshop on Federated Learning: Recent
  Advances and New Challenges}, 2022.

\bibitem{Liu2022}
S.~Liu, Q.~Chen, and L.~You, ``Fed2a: Federated learning mechanism in
  asynchronous and adaptive modes,'' \emph{Electronics}, vol.~11, no.~9, 2022.

\bibitem{Assran_AGPush_2021}
M.~S. Assran and M.~G. Rabbat, ``Asynchronous gradient push,'' \emph{IEEE
  Transactions on Automatic Control}, vol.~66, no.~1, 2021.

\bibitem{Chen2020_ASO_Fed}
Y.~Chen, Y.~Ning, M.~Slawski, and H.~Rangwala, ``{Asynchronous Online Federated
  Learning for Edge Devices with Non-IID Data},'' in \emph{IEEE Big Data 2020},
  Los Alamitos, CA, USA, Dec. 2020.

\bibitem{Wu2022_KAFL}
X.~Wu and C.~Wang, ``{KAFL: Achieving High Training Efficiency for Fast-K
  Asynchronous Federated Learning},'' in \emph{IEEE ICDCS 2022}, Los Alamitos,
  CA, USA, Jul. 2022, pp. 873--883.

\bibitem{wang2022asyncfeded}
Q.~Wang, Q.~Yang, S.~He, Z.~Shi, and J.~Chen, ``Asyncfeded: Asynchronous
  federated learning with euclidean distance based adaptive weight
  aggregation,'' arXiv, 2022, 2205.13797.

\bibitem{Zhang_2023_FedMDS}
Y.~Zhang \emph{et~al.}, ``{FedMDS: An Efficient Model Discrepancy-Aware
  Semi-Asynchronous Clustered Federated Learning Framework},'' \emph{IEEE
  Trans. Parallel Distrib. Syst.}, vol.~34, no.~03, Mar. 2023.

\bibitem{lian2018asynchronous}
X.~Lian, W.~Zhang, C.~Zhang, and J.~Liu, ``Asynchronous decentralized parallel
  stochastic gradient descent,'' arXiv, 2018, 1710.06952.

\bibitem{gossip}
S.~Sundhar~Ram, A.~Nedić, and V.~V. Veeravalli, ``Asynchronous gossip
  algorithms for stochastic optimization,'' in \emph{Proc. of the 48h IEEE
  Conference on Decision and Control (CDC)}, 2009, pp. 3581--3586.

\bibitem{Async-HFL-2023}
X.~Yu \emph{et~al.}, ``{Async-HFL: Efficient and Robust Asynchronous Federated
  Learning in Hierarchical IoT Networks},'' in \emph{Proc. ACM/IEEE IoTDI'23},
  New York, NY, USA, 2023, p. 236–248.

\bibitem{10215505}
Y.~Miao \emph{et~al.}, ``{Robust Asynchronous Federated Learning with
  Time-weighted and Stale Model Aggregation},'' \emph{IEEE Trans Dependable
  Secure Comput}, pp. 1--15, 2023.

\bibitem{ibm_quality_data_metrics}
H.~Patel \emph{et~al.}, ``{Automatic Assessment of Quality of Your Data for
  AI},'' in \emph{5th Joint Int. Conf. on Data Science and Management of Data},
  2022.

\bibitem{10.1145/1541880.1541883}
C.~Batini \emph{et~al.}, ``Methodologies for data quality assessment and
  improvement,'' \emph{ACM Comput. Surv.}, vol.~41, no.~3, Jul. 2009.

\bibitem{cifar10}
A.~Krizhevsky, V.~Nair, and G.~Hinton, ``{The CIFAR-10 dataset},'' 2009.

\bibitem{anderson2018ember}
H.~S. Anderson and P.~Roth, ``{EMBER: An Open Dataset for Training Static PE
  Malware Machine Learning Models},'' arXiv, 2018.

\bibitem{resnet32:2016}
K.~{He}, X.~{Zhang}, S.~{Ren}, and J.~{Sun}, ``{Deep Residual Learning for
  Image Recognition},'' in \emph{Proc. IEEE CVPR 2026}, Las Vegas, NV, USA,
  June 2016, pp. 770--778.

\end{thebibliography}

\vspace{-7ex}
\begin{IEEEbiography}[{\includegraphics[width=1in,height=1.25in,clip,keepaspectratio]{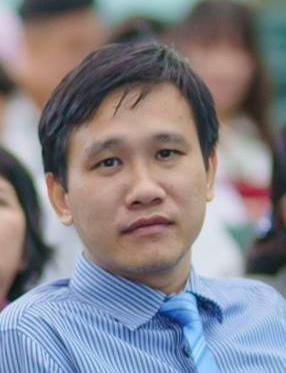}}]{Tien-Dung Cao} received the Ph.D. in Computer Science from the University of Bordeaux, France in 2010. He is currently Dean of the School of Information Technology, Tan Tao University, Vietnam. His research interests include Machine Learning, Big Data, and Service Engineering. He has served as an academic editor of PLOS ONE journal since 2022.
\end{IEEEbiography}
\vspace{-7ex}
\begin{IEEEbiography}[{\includegraphics[width=1in,height=1.25in,clip,keepaspectratio]{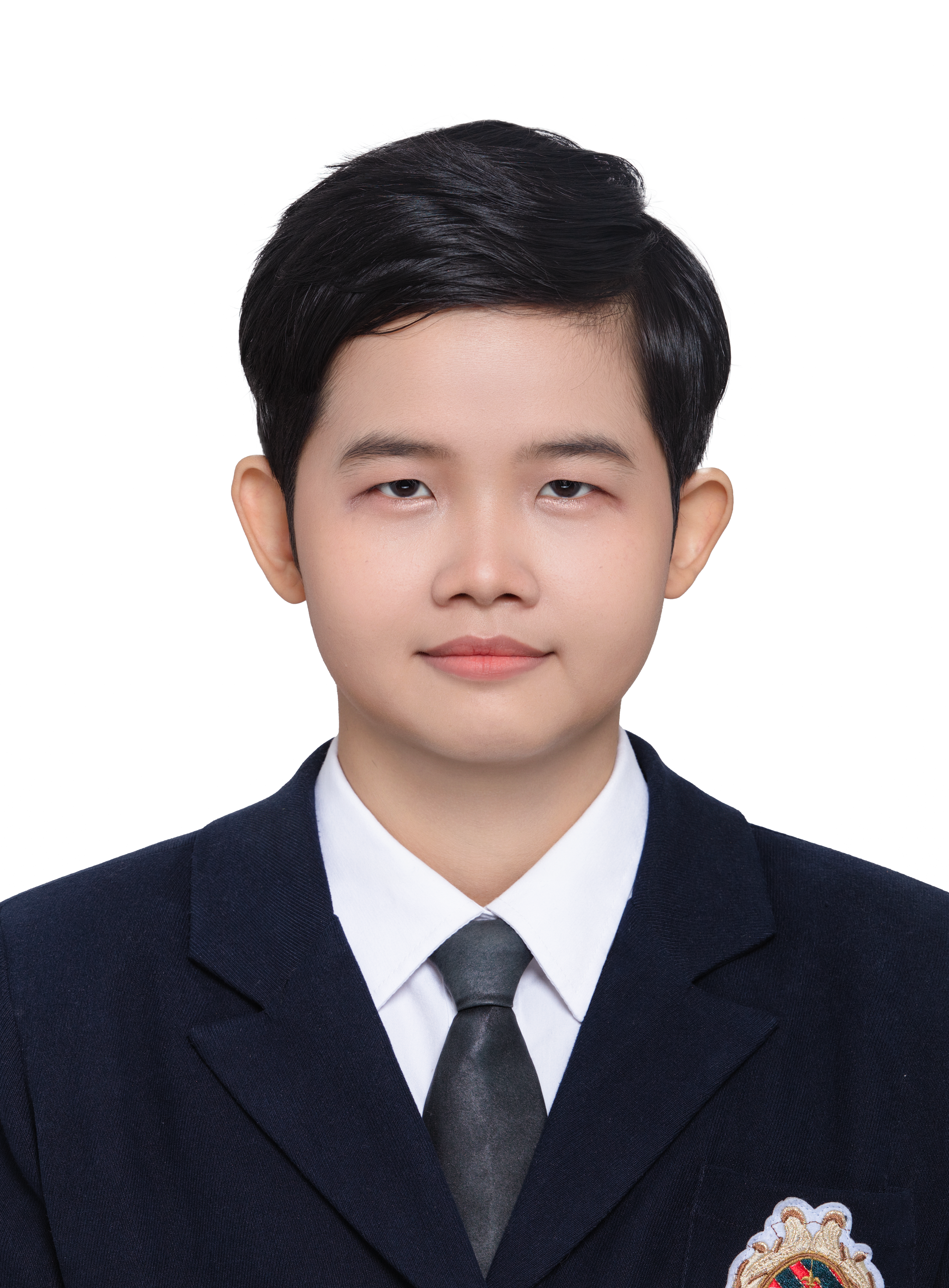}}]{Thao Nguyen Vuong} recently completed a Bachelor's Degree in Computer Science from Tan Tao University. She is now a research intern at Aalto University, Finland. Her current research interests lie in federated learning and automation in machine learning services. She aims to address challenges such as the heterogeneity of training workers in federated systems, including aspects like computational capacity, data distribution, network latency, and availability.
\end{IEEEbiography}
\vspace{-7ex}
\begin{IEEEbiography}[{\includegraphics[width=1in,height=1.25in,clip,keepaspectratio]{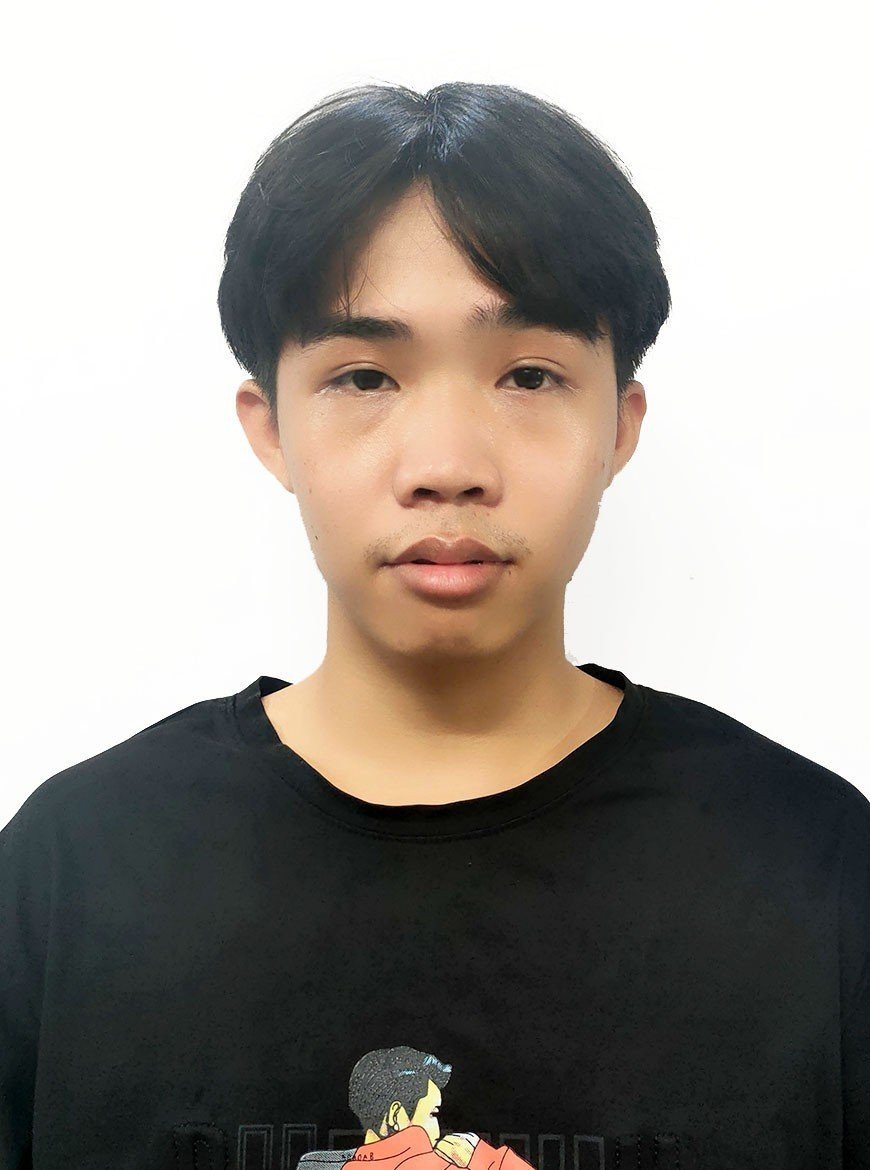}}]{Thai Q. Le} recently completed a Bachelor's Degree in Computer Science from Tan Tao University. He is a passionate programmer who thrives on exploring the uncharted territories of technology. With an insatiable curiosity, he constantly seeks out new and innovative ideas. His research interests include game development which inspires him with new ideas for his work.
\end{IEEEbiography}
\vspace{-7ex}
\begin{IEEEbiography}[{\includegraphics[width=1in,height=1.25in,clip,keepaspectratio]{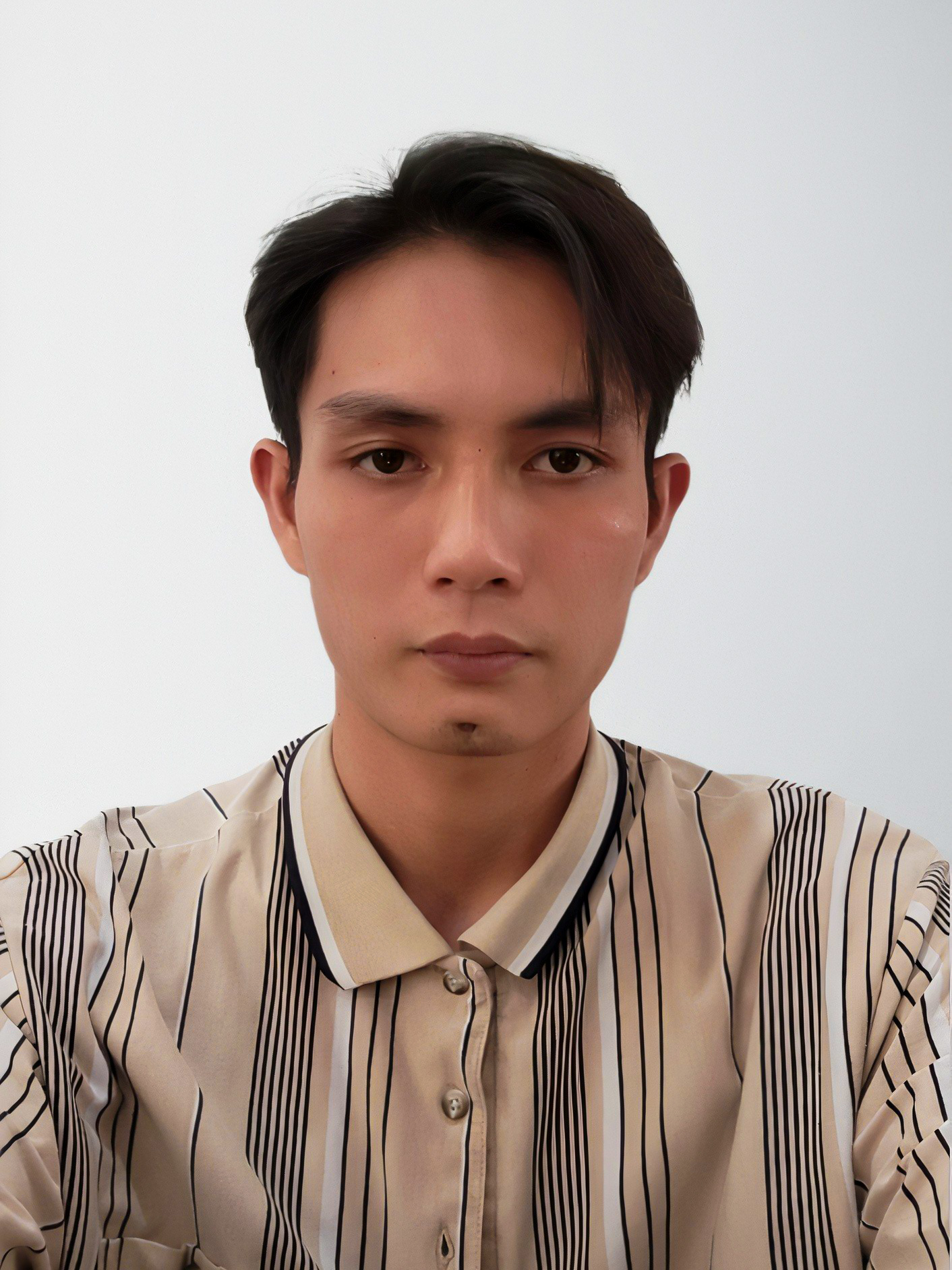}}]{Hoang V. N. Dao} is a software engineer with an insatiable curiosity for technology and a knack for solving complex problems. He recently completed his bachelor's degree in Computer Science with a specialization in AI from Tan Tao University, Vietnam. His research interests include distributed computing, artificial intelligence, and big data.
\end{IEEEbiography}
\vspace{-10ex}
\begin{IEEEbiography}[{\includegraphics[width=1in,height=1.25in,clip,keepaspectratio]{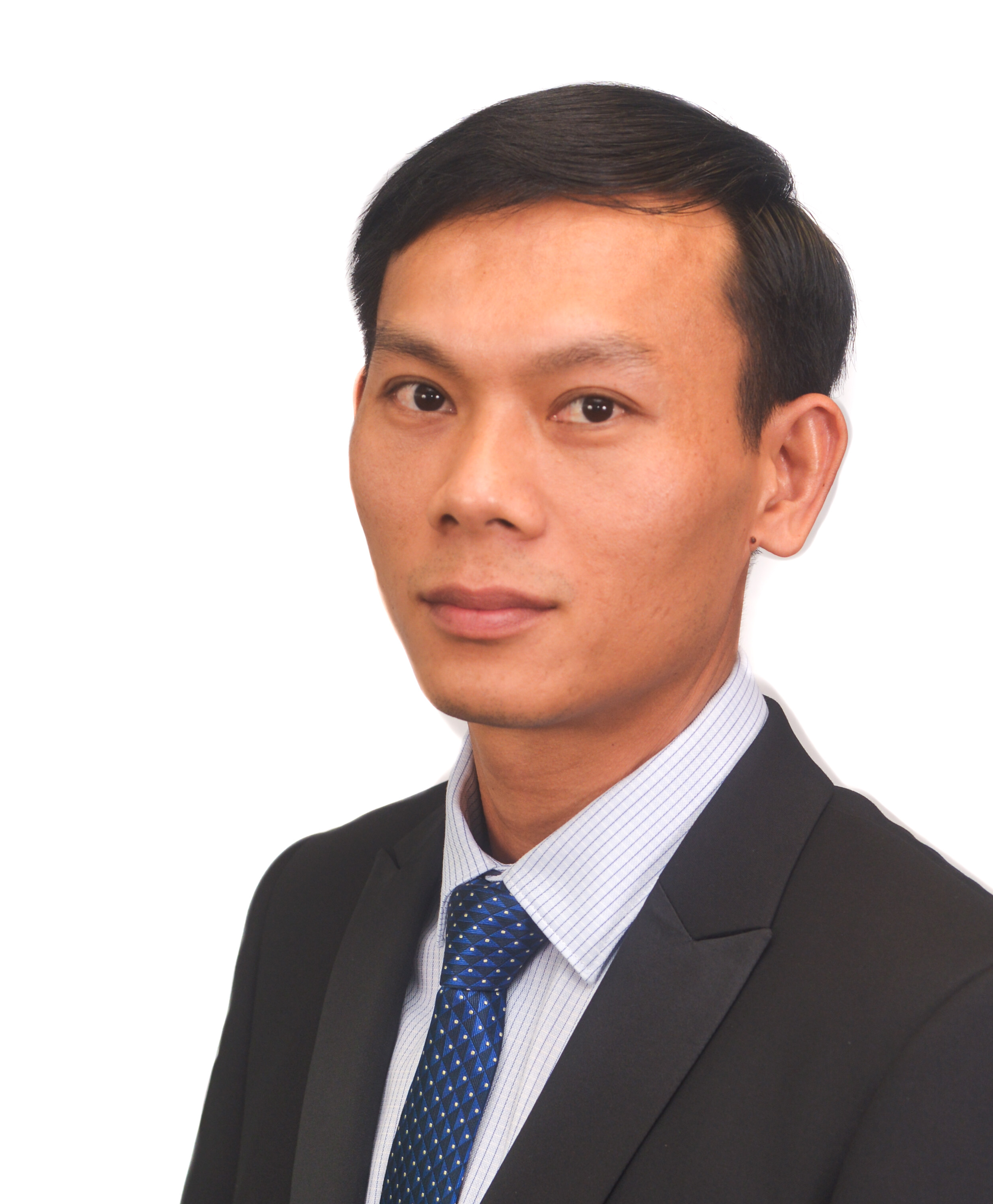}}]{Tram Truong-Huu} is an Assistant Professor at the Singapore Institute of Technology (SIT), Infocomm Technology
(ICT) Cluster. He is currently holding a joint appointment with the Agency for Science, Technology and Research (A*STAR), Singapore, where he has worked as a computer scientist at the Institute for Infocomm Research (I$^2$R) since May 2019. He received the Ph.D. degree in computer science from the University of Nice - Sophia Antipolis (now Côte d’Azur University), France in December 2010. From January 2011 to June 2012, he held a post-doctoral fellowship at the French National Center for Scientific Research (CNRS), France. He worked at the National University of Singapore as a research fellow from July 2012, then senior research fellow from January 2017. His research interests include software-defined networks, the Internet of Things, and the application of artificial intelligence to cybersecurity. He won the Best Presentation Recognition at IEEE/ACM UCC 2013. He has been a member of the IEEE since 2012 and a Senior Member since 2015.
\end{IEEEbiography}

\end{document}